\documentclass[default,iicol]{sn-jnl}% Default with double column layout

\usepackage{graphicx}%
\usepackage{multirow}%
\usepackage{amsmath,amssymb,amsfonts}%
\usepackage{amsthm}%
\usepackage{mathrsfs}%
\usepackage[title]{appendix}%
\usepackage{xcolor}%
\usepackage{textcomp}%
\usepackage{manyfoot}%
\usepackage{booktabs}%
\usepackage{algorithm}%
\usepackage{algorithmicx}%
\usepackage{algpseudocode}%
\usepackage{listings}%

\usepackage{color}
\usepackage{subcaption}
\usepackage{rotating}
\usepackage{array}
\usepackage{gensymb}
\usepackage{makecell}
\usepackage{multirow}
\usepackage{colortbl}

\newcommand{\etal}{\textit{et al}.}
\newcommand{\ie}{\textit{i}.\textit{e}.}
\newcommand{\eg}{\textit{e}.\textit{g}.}
\theoremstyle{thmstyleone}%
\theoremstyle{thmstyletwo}%

\theoremstyle{thmstylethree}%

\begin{document}

\title[Article Title]{Non-aligned Supervision for Real Image Dehazing}

%%=============================================================%%
% Prefix	-> \pfx{Dr}
% GivenName	-> \fnm{Joergen W.}
% Particle	-> \spfx{van der} -> surname prefix
% FamilyName	-> \sur{Ploeg}
% Suffix	-> \sfx{IV}
% NatureName	-> \tanm{Poet Laureate} -> Title after name
% Degrees	-> \dgr{MSc, PhD}
% \author*[1,2]{\pfx{Dr} \fnm{Joergen W.} \spfx{van der} \sur{Ploeg} \sfx{IV} \tanm{Poet Laureate} 
%                 \dgr{MSc, PhD}}\email{iauthor@gmail.com}
%%=============================================================%%

\author[1]{\fnm{Junkai} \sur{Fan}}

\author[1]{\fnm{Fei} \sur{Guo}}
%%\equalcont{These authors contributed equally to this work.}

\author[1]{\fnm{Jianjun} \sur{Qian}}
%%\equalcont{These authors contributed equally to this work.}

\author[2]{\fnm{Xiang} \sur{Li}}
%%\equalcont{These authors contributed equally to this work.}

\author*[1]{\fnm{Jun} \sur{Li}}
%%\equalcont{These authors contributed equally to this work.}

\author*[1]{\fnm{Jian} \sur{Yang}}
%%\equalcont{These authors contributed equally to this work.}

\email{\{junkai.fan, feiguo, csjqian, junli, csjyang\}@njust.edu.cn, xiang.li.implus@nankai.edu.cn}

\affil[1]{\orgdiv{Key Lab of Intelligent Perception and Systems for HighDimensional Information of Ministry of Education \par Jiangsu Key Lab of Image and Video Understanding for Social Security, \par PCA Lab, School of Computer Science and Engineering}, \orgname{Nanjing University of Science and Technology}, \orgaddress{\city{Nanjing}, \country{China}}}

\affil[2]{\orgdiv{College of Computer Science}, \orgname{Nankai University}, \orgaddress{\city{Tianjin}, \country{China}}}

%%==================================%%
%% sample for unstructured abstract %%
%%==================================%%

\abstract{Removing haze from real-world images is challenging due to unpredictable weather conditions, resulting in the misalignment of hazy and clear image pairs. In this paper, we propose an innovative dehazing framework that operates under non-aligned supervision. This framework is grounded in the atmospheric scattering model, and consists of three interconnected networks: dehazing, airlight, and transmission networks. In particular, we explore a non-alignment scenario that a clear reference image, unaligned with the input hazy image, is utilized to supervise the dehazing network. To implement this, we present a multi-scale reference loss that compares the feature representations between the referred image and the dehazed output. Our scenario makes it easier to collect hazy/clear image pairs in real-world environments, even under conditions of misalignment and shift views. To showcase the effectiveness of our scenario, we have collected a new hazy dataset including 415 image pairs captured by mobile Phone in both rural and urban areas, called "Phone-Hazy". Furthermore, we introduce a self-attention network based on mean and variance for modeling real infinite airlight, using the dark channel prior as positional guidance. Additionally, a channel attention network is employed to estimate the three-channel transmission. Experimental results demonstrate the superior performance of our framework over existing state-of-the-art techniques in the real-world image dehazing task. Phone-Hazy and code will be available at
\href{https://fanjunkai1.github.io/projectpage/NSDNet/index.html}{https://fanjunkai1.github.io/projectpage/NSDNet/index.html}.}


%%\pacs[JEL Classification]{D8, H51}

%%\pacs[MSC Classification]{35A01, 65L10, 65L12, 65L20, 65L70}

\maketitle

\begin{figure*}[ht]
	\centering
	\includegraphics[width=0.96\linewidth]{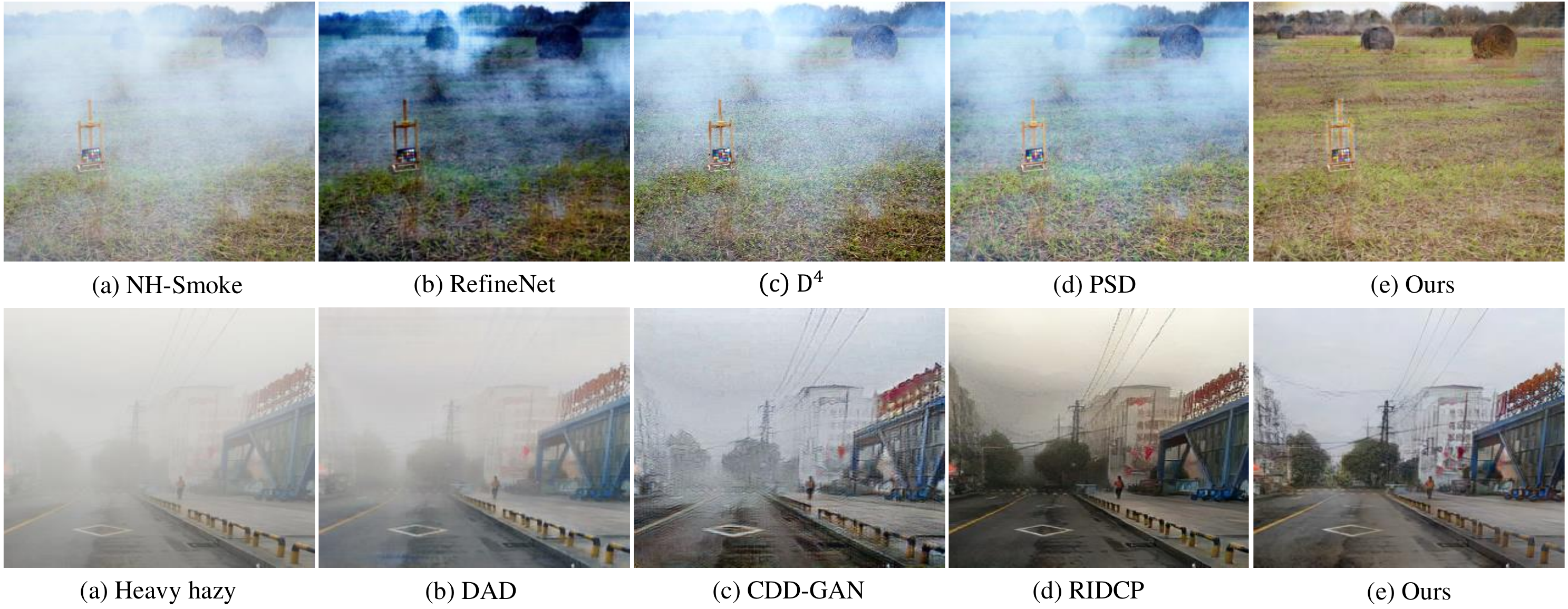}
	\vskip 0.05in
	\caption{Dehazing results on real-world images from CVPRws 2021~\cite{jo2021multi} and our Phone-Hazy. Our method can generate much clearer results compared to the state-of-the-art methods, RefineNet~\cite{zhao2021refinednet}, D$^{4}$~\cite{yang2022self}, PSD~\cite{chen2021psd}, DAD~\cite{shao2020domain}, CDD-GAN~\cite{chen2022unpaired} and RIDCP~\cite{wu2023ridcp}.} %The non-homogeneous smoke and dense hazy images are selected from CVPR-workshop 2021 \cite{jo2021multi} and RTTS \cite{li2018benchmarking}, respectively.
	\label{fig:1}
	%\vskip -0.1in
\end{figure*}%

\section{Introduction}\label{sec1}

\label{sec:intro}
Hazy is a main restrictive factor to capturing a degraded image (\eg, limited visibility, low contrast) during the imaging formation process, resulting in the poor performance of many computer vision applications, such as object detection~\cite{Hahner_2021_ICCV}, scene recognition~\cite{bijelic2020seeing}, depth estimation~\cite{wang2021regularizing}, semantic segmentation~\cite{sakaridis2018model} and autonomous driving~\cite{codevilla2019exploring}. Therefore, image dehazing, aiming to restore a clear image from a hazy input, has received more and more attention in the past decade~\cite{dong2020multi,chen2021psd,wu2021contrastive,zheng2021ultra}. The hazing process is typically modeled as an atmospheric scattering model:
\begin{equation}
I(x) = J(x)t(x)+A_{\infty}(\lambda)(1-t(x)),
\label{eq:scatter}
\end{equation}
where $x$ represents a pixel position, $I(x)$ is an input hazy image, $J(x)$ is a restored scene radiance, $A_{\infty}$ is an infinite airlight (the airlight radiance corresponding to an object at an infinite distance, \eg, horizon), $t(x)=e^{-{\beta(\lambda)}d(x)}$ is a transmission map, $d(x)$ and $\beta(\lambda)$ denote a scene depth and a scattering coefficient with a wavelength of light $\lambda$, respectively.
%However, most of the current dehazing work is carried out based on this model. ???however????? % physical priors \cite{}, color attenuation prior \cite{zhu2015fast} (\eg, maximum contrast prior \cite{tan2008visibility}, dark channel prior \cite{he2010single} and color ellipsoid prior \cite{bui2017single})
%Following this atmosphere scattering model, most of the existing deep learning methods can be divided roughly into two categories: direct dehazing and indirect dehazing. The former uses the aligned hazy/clear image pairs to train an end-to-end deep dehazing network in supervised setting for directly predicting the clear image. The latter conducts popular CNN-based networks (\eg, DCPDN \cite{zhang2018densely} and BidNet \cite{pang2020bidnet}) to indirectly estimate $A_{\infty}$ and $t(x)$ for restoring the clear image via the scattering model. Although the above methods have shown good performance, they still suffer from two challenges as follows.
Following the atmosphere scattering model, most deep learning methods~\cite{yang2018towards,li2019heavy,zhang2018densely,pang2020bidnet,zhao2021refinednet,li2021you} utilize widely-adopted CNN-based networks~\cite{he2016deep,ronneberger2015u} to construct dehazing, airlight, transmission networks to estimate $J$, $A_{\infty}$ and $t$ from the input hazy image $I$, as outlined in Eq.~\eqref{eq:scatter}. These methods can be broadly categorized into paired and unpaired approaches, taking into consideration the relationship between hazy and clear image pairs. However, despite their promising results, these methods still encounter three significant challenges.

% and the transmission map
%For the direct dehazing methods, firstly, it is practically impossible to collect a large number of the aligned hazy/clear image pairs from the real scenes due to complex and dynamic environments, easily leading to pixel misalignment and shifted views. This leads to poor restorations on real-world hazy images using the dehazing models as they have no choice but to be trained on synthetic data. Some recent works \cite{shao2020domain} try to use domain adaptation to reduce the significant gap between synthetic and real data. However, they still face the unsatisfied result as the data are sampled from the different distributions (scenes). For the indirect dehazing methods, secondly, the popular CNN-based networks are not well designed for the real infinite airlight since they cannot accurately model its variations determined by the wavelength of light and the size of the scattering particles in real-world \cite{mccartney1976optics} and lack useful priors to create their network structures, resulting in an inaccurate estimation and an unsatisfied image restoration.

Firstly, the majority of the paired methods~\cite{shao2020domain,yu2022source,li2022physically,ye2022mutual} employ the aligned hazy/clear image pairs within a supervised setting to train the dehazing network, and then restore the real hazy image through domain adaptation. The aligned image pair used for training are often synthetically generated, utilizing the atmospheric scattering model described in Eq.~\eqref{eq:scatter} to convert clear images into hazy counterparts. Nonetheless, this yields subpar results when applied to real-world hazy images due to the inherent divergence between synthetic and authentic hazy image domains, such as the DAD model (\cite{shao2020domain}) in Fig.~\ref{fig:1}.
Moreover, it is unfeasible to collect a large number of perfectly aligned hazy/clear image pairs from real-world scenes. This is primarily because these images are typically captured under different times, weather conditions and camera perspectives, resulting in pixel misalignment and shift views.

Secondly, the unpaired methods~\cite{yang2018towards,zhao2021refinednet,chen2022unpaired,yang2022self}rely on unpaired sets of clear/hazy images for training dehazing networks. While gathering unpaired images is relatively uncomplicated, they are drawn from distinct distributions or scenes. As a result, training becomes challenging, and the dehazing outcomes suffer, as evident in the performance of models like RefineNet (\cite{zhao2021refinednet}), CDD-GAN (\cite{chen2022unpaired}), and D$^{4}$ (\cite{yang2022self}), showcased in Fig.~\ref{fig:1}.

Thirdly, the above mentioned methods usually assume a constant value for the airlight $A_{\infty}$. However, $A_{\infty}$ varies due to differences in the size of scattering particles and the wavelength of light (\cite{mccartney1976optics, zhou2021learning}) in real-world scenarios. Consequently, a fixed $A_{\infty}$ fails to capture these variations, resulting in unsatisfactory dehazing outcomes.

To address these problems, we develop a non-aligned supervision framework, which consists of dehazing, infinite airlight, and transmission networks rooted on the atmosphere scattering model. One important idea is the utilization of non-aligned clear images to supervise the dehazing network. This allows for clear images that are not perfectly aligned with the hazy images to be used for training, producing two valuable benefits. Unlike the paired methods, our approach not only relaxes the strict alignment constraint, but also becomes easier to collect non-aligned image pairs from the same scene under more relaxed conditions. In contrast to the unpaired methods, our approach reduce the difference between hazy and clear image distributions, making the model easy learning. Furthermore, we introduce a multi-scale reference loss that combines adversarial and contextual losses to optimize the dehazing network using multi-scale non-aligned pairs. 

\begin{figure*}[t]
\centering
\includegraphics[width=0.96\linewidth]{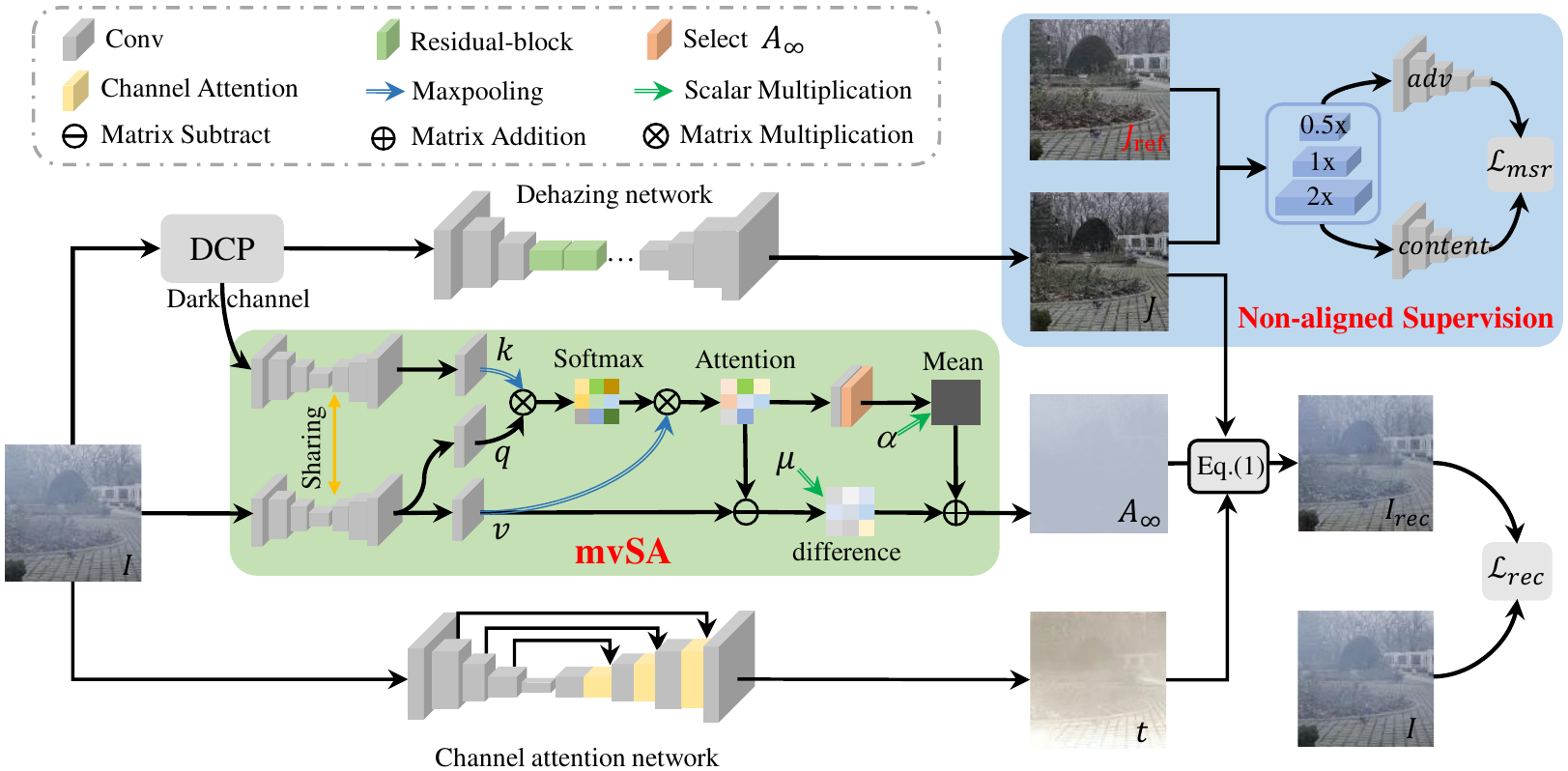}
\caption{Overall pipeline of our non-aligned supervision framework grounded in the atmosphere scattering model for the real image dehazing. This framework comprises essential components: a generator network for the dehazing image $J$, a mvSA network for the airlight map $A_{\infty}$, and a channel attention network for the transmission map $t$. Our heart part is the non-aligned supervision setting to train the dehazing generator network by leveraging a clear and non-aligned reference image, conveniently obtainable from the same scene. Another important part is the mvSA network to effectively estimate the $A_{\infty}$ by using dark channel prior in real scenes. Note that our framework stands apart from conventional supervised dehazing models as it operates without the need for the aligned ground truths.}  %The framework consists of the mvSA module and the non-aligned supervision module.} 
\label{fig:nasframework}
%\vskip -0.12in
\end{figure*}

Another perspective involves regarding $A_{\infty}$ as a non-uniform mapping. To more effectively model the $A_{\infty}$ within the hazy images, we present a mean and variance self-attention (mvSA) network by incorporating both the dark channel prior (DCP)~\cite{he2010single} and the hazy image. mvSA is able to highlight hazy features by capturing their average values and predicting the fluctuations caused by factors such as wavelength effects and scattering particles. In comparison to both DCP and traditional self-attention (SA), our mvSA network offers a more precise constraint on the range of infinite airlight. Overall, our contributions are summarized as follow:

% (\ie, similar to the $A_{\infty}$ of DCP)
% 1) we construct a self-attention operation to highlight hazy features by fusing DCP and the hazy image to compute its mean (\ie, similar to the $A_{\infty}$ of DCP). 2) we add an output head to predict its variations caused by the effects of wavelength and scattering particles. Compared with DCP and self-attention (SA), mvSA more accurately constrains the range of infinite airlight.
%In addition, we employ a channel attention network and a guide filter \cite{he2010single} to estimate the refined three-channel transmission map, which are better than the one-channel transmission map of the conventional method \cite{li2019heavy,yang2018towards}.
% by combining the atmospheric scattering model and CNN
\begin{itemize}
\item We propose a novel non-aligned supervision framework for effectively dehazing real-world images, as depicted in Fig.~\ref{fig:nasframework}. By incorporating a clear and non-aligned reference image for dehazing network supervision, we effectively alleviate the stringent alignment prerequisites typically associated with hazy/clear image pairs. As far as we know, we are pioneers in exploring non-aligned supervision for single image dehazing in real-world scenarios.
\item We present a mean and variance self-attention network (mvSA) that utilizes the dark channel prior as position guidance to better model the infinite airlight. Our experimental results also confirm its effectiveness.
\item We provide a real-world hazy dataset including 415 non-aligned hazy/clear image pairs, which are collected manually by using iPhone XR in the different real scenes (\ie, countryside and urban roads). 
\end{itemize}

\section{Related work}
%{\bf Single image dehazing.} ????????????

Here, we mainly review two categories: prior-based dehazing methods and learning-based dehazing methods.

\textbf{Prior-based methods} for haze removal rely on atmospheric scattering theory~\cite{mccartney1976optics} and employ manually crafted priors based on empirical observations. These priors mainly include contrast maximization~\cite{tan2008visibility}, dark channel prior (DCP)~\cite{he2010single}, color attenuation prior~\cite{zhu2015fast}, color-line prior~\cite{fattal2014dehazing}, and non-local prior~\cite{berman2016non}. Although effective in general, these methods may yield suboptimal results in complex real-world scenarios, particularly in sky regions where DCP struggles due to its reliance on priors that are specific to certain scenes and assumptions.

{\bf Learning-based methods} use deep neural networks to learn either the parameters (\ie, infinite airlight $A_{\infty}$, transmission map $t(x)$ and hazy-free $J(x)$) of the atmospheric scattering model or the mapping between hazy and clear images. For the former, many works~\cite{cai2016dehazenet,ren2016single,mondal2018image,zhang2018densely,zhang2019joint,li2019heavy,li2020zero,pang2020bidnet,park2020fusion,li2021you,liu2022towards} focused on directly estimating $t(x)$ and $A_{\infty}$ by using CNNs or GANs. Mondal \etal.~\cite{mondal2018image} proposed that airlight is non-uniform and defined it as $A_{\infty}(\lambda)(1-t(x))$, while our infinite airlight only refers to $A_{\infty}$. Due to the difficulty of obtaining ground truths for $t(x)$ and $A_{\infty}$ in real scenes, these methods are difficult to achieve good results. To improve the accuracy of $A_{\infty}$ and $t(x)$, some methods~\cite{liu2019learning,li2019lap,wu2019accurate,kar2020transmission} used an iterative optimization approach to obtain their optimal values, but with limited gains. The latter primarily relies on designing various network modules to efficiently extract features, without utilizing the atmospheric scattering model. Most of them \cite{li2018single,qu2019enhanced,qin2020ffa,cong2020discrete,deng2020hardgan,dong2020fd,shyam2021towards} construct dehazing models using GAN variants as a basis and introduce perceptual loss~\cite{johnson2016perceptual} as a constraint. Part of these GAN variants mainly utilize multi-scale and attention mechanismes(\eg, channel attention, spatial attention) to efficiently extract hazy features, such as~\cite{qu2019enhanced, qin2020ffa}. Besides, similar network architecture design ideas have also appeared in CNN-based dehazing networks~\cite{li2017aod,zhang2018densely,ren2018gated,deng2019deep,liu2019griddehazenet,dong2020multi,ye2021perceiving}. Recently, visual transformers (ViT) is used to design different structures for improving dehazing performance~\cite{zhao2021hybrid,song2022vision,valanarasu2022transweather,guo2022image}. The aforementioned works mainly employ supervised learning methods that heavily rely on ground truth data, and are primarily trained on synthetic hazy (\ie, SOTS~\cite{li2018benchmarking}, Cityscapes \cite{cordts2016cityscapes}). However, the dehazing effect is not ideal due to limited depth and degraded image quality, resulting in domain gaps, and a lack of real hazy/clear image pairs.

To address this issue, some works have proposed domain adaptive and unpaired dehazing models for real-world. These models are mainly built on the framework of CycleGAN~\cite{zhu2017unpaired}, such as Cycle-Dehaze~\cite{engin2018cycle}, DAD~\cite{shao2020domain}, and D$^{4}$. Chen \etal.~\cite{chen2021psd} proposed a dehazing framework pre-trained on synthetic datasets and fine-tuned on real data with physical priors. Similarly, Wu \etal.~\cite{wu2023ridcp} enhanced post-fine-tuning image quality by using high-quality codebook priors. Yang \etal.~\cite{yang2018towards} employed deep networks to estimate $A_{\infty}$, $t(x)$, and $J(x)$ separately for reconstructing hazy images, and constraints dehazing results by using unpaired hazy/clear images, similar works include~\cite{zhao2021refinednet}. Both of these approaches do not perform well in real-world scenarios, mainly because of differences in sample distributions. Compared to the domain adaptive and unpaired supervision, our non-aligned supervision is a stronger constraint.

Different from the above mentioned methods, Our method outperforms previous dehazing models by training on paired real hazy datasets and extracting effective features from non-aligned reference images. Furthermore, we have redefined the $A_{\infty}$ (\ie, non-uniform map) and proposed a novel network (mvSA) that more accurately estimates $A_{\infty}$ in real-world scenarios.

\section{Methodology}

In this section, we propose a \textbf{N}on-aligned \textbf{S}upervision \textbf{D}ehazing framework by constructing three dehazing, airlight, and transmission \textbf{Net}works (called \textbf{NSDNet}) from the input hazy image, as shown in Fig.~\ref{fig:nasframework}.  Our key idea is to explore a non-aligned supervision setting that the dehazing network is supervisedly trained by using a clear and non-aligned reference image. Another idea is to construct a mean and variance self-attention (mvSA) network for predicting better airlight $A_{\infty}$ by using dark channel prior \cite{he2010single}. Before showing them, we first give the dehazing and transmission networks.

\emph{The dehazing network} is designed to directly generate a haze-free image from the input hazy image. As depicted in Fig.~\ref{fig:nasframework}, we compute a rough haze-free image using the DCP method~\cite{he2010single}. It is put into the dehazing network, which is a generator network of CycleGAN~\cite{zhu2017unpaired}.

\emph{The transmission network} aims to produce a three-channel transmission map through the utilization of a channel attention network from the input hazy image, as demonstrated in Fig.~\ref{fig:nasframework}. Its architecture is an encoder-decoder structure with skip connection across the feature scales~\cite{ronneberger2015u}. Ultimately, the guided filter~\cite{he2010single} is employed to derive the final transmission map.

%Different from the definition of transmission map in previous work \cite{he2010single}. In this work, we define transmission map to be a three-channel variable , and present a channel attention network to estimate the transmission map. 

We denote $I\in \mathbb{R}^{3 \times H \times W}$ by the input hazy image, $J_{\text{ref}}\in \mathbb{R}^{3 \times H \times W}$ by the clear and
non-aligned reference image, $J\in \mathbb{R}^{3 \times H \times W}$ by the output of the dehazing network with the input $I$, $t \in \mathbb{R}^{3 \times H \times W}$ by the output of the transmission network. Note that $J_{\text{ref}}$ is not aligned to $I$ or $J$.

\subsection{Non-aligned Supervision}
To effectively address both the reduction of the domain gap between synthetic and real hazy images in the paired methods, and the minimization of disparities between hazy and clear image distributions in the unpaired methods, a logical approach is to amass non-aligned hazy/clear image pairs ($I$, $J_{\text{ref}}$) within the context of the same real-world scenes. These pairs can then serve as supervisory signals for guiding the dehazing network's training. The detailed strategy for assembling our Phone-hazy dataset is outlined in Appendix \ref{Appendix_A:phone-hazydataset}. In this subsection, we establish the concept of non-aligned supervision as a reference loss between $J_{\text{ref}}$ and $J$, determined by assessing their feature similarity. This assessment is conducted using multi-scale augmentation, as described below.

\textbf{Multi-scale reference loss} encompasses both a multi-scale adversarial loss and a multi-scale contextual loss, comparing $J_{\text{ref}}$ and $J$. The augmentation of both $J_{\text{ref}}$ and $J$ employs three identical scales: $0.5\times$, $1\times$, and $2\times$, which are represented in this work as $\mathbf{J}_{\text{ref}}=\{J_{\text{ref}}^i\}_{i=1,2,3}$ and $\mathbf{J}=\{J^i\}_{i=1,2,3}$. Mathematically, the multi-scale reference loss can be expressed as follows:
\begin{align}
\mathcal{L}_{\text{msr}}\left(\mathbf{J}_{\text{ref}}, \mathbf{J}\right) =& \omega_{1}  \mathcal{L}_{\text{msa}}\left(\mathbf{J}_{\text{ref}}, \mathbf{J}\right) \nonumber\\
&+\omega_{2}\mathcal{L}_{\text{msc}}\left(\mathbf{J}_{\text{ref}}, \mathbf{J}\right),
\label{eq:msr}
\end{align}
where $\omega_{1}$ and $\omega_{2}$ are positive hyper-parameters to balance the two losses. They are set to 1 by default in this paper.

\emph{\textbf{Multi-scale adversarial loss.}} Given the potent capability of adversarial learning~\cite{goodfellow2014generative} to generate highly realistic images, our approach leverages adversarial loss to supervise both the dehazing network (generator) and a discriminator network. The discriminator network, a compact CNN comprising 5 layers, utilizes convolution to yield a one-dimensional output. All layers incorporate 4 $\times$ 4 Convolution-BatchNorm-LeakyReLU units with a stride of 2, except for the first and last layers, which lack BatchNorm. Furthermore, we expand the adversarial loss into a multi-scale variant, expressed as follows:
\begin{align}
\mathcal{L}_{\text{msa}}\left(\mathbf{J}_{\text{ref}}, \mathbf{J}\right) %=& \sum_{i=1}^3\mathcal{L}_{\text{ms-adv}}\left(J_{\text{ref}}^i,J^i\right) \notag\\
=&\sum_{i=1}^3\Big(\mathbb{E}_{J_{\text{ref}}^i}\left[ \log D(J_{\text{ref}}^i)\right] \notag\\
	& + \mathbb{E}_{J^i}\left[ \log (1 - D(J^i))\right]\Big),
\end{align} 
where the index $i$ represents the different scales, and $J$ is the output of the dehazing network through training with the above loss.

\emph{\textbf{Multi-scale contextual loss.}}
To better explore the clear and unaligned reference image, we draw inspiration from the contextual loss~\cite{mechrez2018contextual}. This loss quantifies cosine similarity distances between unaligned images, initially designed for image-to-image translation tasks. Our contribution is the extension to a multi-scale contextual loss, defined as follows:
\begin{align}
\mathcal{L}_{\text{msc}}\left(\mathbf{J}_{\text{ref}}, \mathbf{J}\right) %= & \sum_{i=1}^3\mathcal{L}_{\text{ms-con}}\left(J_{\text{ref}}^i,J^i\right) \notag\\
=& \sum_{i=1}^3 \sum_{l=1}^3 -\log\left[S\left(\varPhi^{l}(J^i),
\varPhi^{l}(J_{\text{ref}}^i)\right)\right],
\end{align}
where $S$ denotes the contextual similarity between image features, and $\varPhi^{l}(J)$ as well as $\varPhi^{l}(J_{ref})$ refer to feature maps extracted from layer $l$ of the VGG-16 network $\varPhi$ using inputs $J$ and $J_{ref}$, respectively.

\textbf{Remark 1.} Our approach to non-aligned supervision offers three significant advantages. Firstly, it alleviates the stringent alignment requirements, particularly when reducing the domain gap between synthetic hazy/clear image pairs in paired methods. Secondly, it strengthens the coherence between hazy and clear image distributions within unpaired methods. Thirdly, it facilitates the collection of non-aligned hazy/clear image pairs from authentic scenes, even when subjected to relaxed conditions like pixel misalignment and varying viewpoints. Additionally, we have leveraged the iPhone XR to curate a real-world hazy dataset, referred to as Phone-Hazy.

\subsection{Mean and Variance Self-attention}
In this subsection, we introduce a mean and variance self-attention (mvSA) network, designed to enhance modeling of the infinite airlight denoted as $A_{\infty}$. By incorporating the informative DCP prior~\cite{he2010single}, we can find the high-quality infinite airlight from an input hazy image $I$. To compute the dark channel map $D$ of the input image $I$, we adopt the DCP method~\cite{he2010single}. Subsequently, leveraging a shared network, we extract features from both the dark channel map $D$ and the hazy image $I$, which are then input into our mvSA network for more accurate estimation of $A_{\infty}$, as depicted in Fig.~\ref{fig:nasframework}.

{\bf The shared network} is an encoder-decoder structure with skip connection across the feature scales~\cite{ronneberger2015u}. This network is utilized to extract features from the dark channel image $D$ and the haze image $I$, resulting in feature representations denoted as $F_{d} \in \mathbb{R}^{B \times C \times H \times W}$ and $F_{h} \in \mathbb{R}^{B \times C \times H \times W}$ respectively. Here, $B$, $C$, $H$, and $W$ represent the batch size, number of channels, height, and width of the feature tensors. \textit{Note that more details are provided in supplementary material.} % Firstly, the image go through four down-sampling blocks, and each down-sampling block consists of one 4$\times$4 convolutional layer with stride 2 and two 3$\times$3 convolutional layers with stride 1, then ReLU and BN are followed by the convolutional layer. The output channels of four down-sampled layers are setting as 64, 128, 256, and 512, respectively.  Concatenations are then employed with features across scales corresponding to the same dimension. Through the top-down and bottom-up structure, the obtained dark channel feature and haze feature are operated on by self-attention.

{\bf The mvSA network} utilizes mean and variance self-attention mechanisms to estimate the infinite airlight denoted as $A_{\infty}$. This estimation is achieved through the utilization of the dark channel prior, which serves as positional guidance. Employing the principles of self-attention~\cite{wang2018non}, the network accentuates the hazy regions within both the dark channel feature $F_{d}$ and the hazy feature $F_{h}$. The mean value of the top 1\% brightest pixels within the hazy region is adopted as the representative mean for $A_{\infty}$. Furthermore, the network formulates the disparity between the hazy feature and the self-attention feature as a predictor for the variance of $A_{\infty}$. The mathematical depiction of the mvSA network is presented as follows.

%To make the network to better learn the features of haze regions, inspired by non-local neural networks \cite{wang2018non}, we use self-attention to highlight the similar regions (i.e. the haze region.) of the dark channel feature and the haze feature. Then to take the mean of top 1\% brightest pixels as relative mean by select $A_{\infty}$ block. Besides, the difference between the haze feature and the self-attention feature map is used as the perturbation term (i.e., the effect of wavelength of $A_{\infty}$  ) of estimation $A_{\infty}$ , called relative variance, the take sum of relative mean and relative variance  as $A_{\infty}$. Fig. \ref{fig:1} shows the structure of $A_{\infty}$ attention module. 

Following the extraction of features $F_{d}$ and $F_{h}$ from the aforementioned shared network, we employ convolutional operations with a kernel size of $1\times1$ to transform these features, yielding the embedding vectors $q_{h}$, $k_{d}$, and $v_{h}$. These transformations are denoted as $C_{1\times1}^{q}$, $C_{1\times1}^{k}$, and $C_{1\times1}^{v}$, respectively. The resulting embeddings can be represented as:
\begin{align}
q_{h} = C_{1\times1}^{q}(F_{h}),\ k_{d} = C_{1\times1}^{k}(F_{d}), \nonumber \\ v_{h} = C_{1\times1}^{v}(F_{h}), \ \ \ \ \ \ \ \ \ 
	\label{eq:kqv}
\end{align}where $q_{h} \in \mathbb{R}^{B \times \frac{C}{8} \times H \times W}$, $k_{d} \in \mathbb{R}^{B \times \frac{C}{8} \times H \times W}$ and $v_{h} \in \mathbb{R}^{B \times \frac{C}{8} \times H \times W}$.
To manage computational complexity, we perform a downsampling operation on $k_{d}$ and $v_{h}$ using a $4\times4$ maxpooling operation denoted as $\mathcal{M}_{4\times4}$. These downsampled versions are defined as:
\begin{align}
\widehat{k}_{d} = \mathcal{M}_{4\times4}(k_{d}),\ \ \ \widehat{v}_{h} = \mathcal{M}_{4\times4}(v_{h}),
\label{eq:kvreduce}
\end{align}where $\widehat{k}_{d} \in \mathbb{R}^{B \times \frac{C}{8} \times \frac{H}{4} \times \frac{W}{4}}$ and $\widehat{v}_{h} \in \mathbb{R}^{B \times \frac{C}{8} \times \frac{H}{4} \times \frac{W}{4}}$. The attention weight is computed through matrix multiplication between reshaped $q_{h} \in \mathbb{R}^{B\times\frac{C}{8}\times(HW)}$ and reshaped $\widehat{k}_{d} \in \mathbb{R}^{B \times \frac{C}{8} \times \frac{HW}{16}}$, followed by applying the softmax activation. Subsequently, the attention map $F_{\text{att}} \in \mathbb{R}^{B \times \frac{C}{8} \times HW}$ is obtained by multiplying the attention weight with reshaped $\widehat{v}_{h} \in \mathbb{R}^{B\times \frac{C}{8}\times \frac{HW}{16}}$, which is written as:
\begin{equation}
	F_{\text{att}} = \text{softmax}(q_{h}^{T} \otimes \widehat{k}_{d}) \otimes \widehat{v}_{h}^{T},
\end{equation}where $\otimes$ denotes matrix multiplication and $F_{\text{att}} \in \mathbb{R}^{B\times(HW)\times \frac{C}{8} }$. Utilizing the reshaped attention map $F_{\text{att}} \in \mathbb{R}^{B \times \frac{C}{8} \times H \times W}$ and the embedding $v_{h}$, we calculate the mean and variance of the infinite airlight $A_{\infty} \in \mathbb{R}^{B \times 3 \times H \times W}$ as follows:
\begin{align}
	A_{\infty} &= \alpha A_{m} + \mu A_{v}, \label{eq.5}\\
	A_{v}  &= C_{1 \times 1 }(|v_{h} - F_{\text{att}}|), \label{eq.6}\\
	A_{m} &= U_{A_{\infty}}\left [C_{1 \times 1 }(F_{\text{att}}) \right ],
	\label{eq.7}
\end{align}
\noindent where $A_{m} \in \mathbb{R}^{B \times 3 \times H \times W}$  and  $A_{v} \in \mathbb{R}^{B \times 3 \times H \times W}$ denote the relative mean and relative variation, respectively. The terms $\alpha$ and $\mu$ serve as adjustment factors for their corresponding components. The operation $C_{1 \times 1}(\cdot)$ represents a convolutional operation with a $1\times1$ filter to reduce the number of channels. The notation $U_{A_{\infty}}[\cdot]$ indicates the selection of the top 1\% brightest pixels within $A_{\infty}$ from the attention feature map.

\textbf{Remark 2.} Our mvSA network presents a superior and more encompassing approach for estimating the mean and variance maps of infinite airlight within real scenes. This outperforms previous studies~\cite{he2010single,zhang2018densely,li2019heavy,liu2019learning,zhao2021refinednet,chen2021psd}, which solely account for limited three-channel constants and consequently fall short in capturing the inherent variation. %Furthermore, the variation learned by our mvSA network is different from the simple Gaussian assumption \cite{sun2020conditional}.

%%%%%%%%%%%%%%%%%%%%%%%%%%%%%%%%%%%%%%%%%%%%%%%%%%%%%%%%%%%%%%%%%%%%%%%%%%%%%%
% \scriptsize \footnotesize \small
\begin{table*}[ht]
	\linespread{1.0}
	\centering
	\setlength{\tabcolsep}{2pt}
	\renewcommand{\arraystretch}{1.4}
	\resizebox{0.98\linewidth}{!}{
	\begin{tabular}{c|c|cccc|cc|ccc|c|c} 
		\Xhline{1.3pt} 
		\multicolumn{1}{c|}{\multirow{2}[1]{*}{\makecell{Data\\Settings}}}& \multicolumn{1}{c|}{\multirow{2}[1]{*}{Methods}} & \multicolumn{4}{c|}{Real-world Smoke} & \multicolumn{2}{c|}{Phone-Hazy}  & \multicolumn{3}{c|}{RTTS (Only testing)} &\multirow{2}[1]{*}{\makecell{Params\\(M)}} & \multirow{2}[1]{*}{Ref.} \\
		%\cline{3-11}
		&     & \multicolumn{1}{c}{PSNR$\uparrow$} & \multicolumn{1}{c}{SSIM$\uparrow$} & \multicolumn{1}{c}{FADE$\downarrow$} & \multicolumn{1}{c|}{NIQE$\downarrow$} & \multicolumn{1}{c}{FADE$\downarrow$} & \multicolumn{1}{c|}{NIQE$\downarrow$} & \multicolumn{1}{c}{FADE$\downarrow$} & \multicolumn{1}{c}{NIQE$\downarrow$} & \multicolumn{1}{c|}{Votes$\uparrow$} & \\
		\Xhline{0.7pt} 
		 \multirow{5}{*}{Unpaired}  
		& DCP \cite{he2016deep} & 15.01 & 0.39 & 0.3961 & 4.1303 & 0.8235  & 5.4222 & 0.7846 & 6.0006 & - &- & \small CVPR09 \\
		& DisentGAN \cite{yang2018towards} & 14.93 & 0.27 & 0.7081 & 5.5187 & 0.9235 & 5.1244 & 1.2620 & 6.1112 & - & 11.48 & \small{AAAI18} \\
		& RefineNet \cite{zhao2021refinednet} & 15.44 & 0.42 & 0.3042 & 4.1840  & 1.0889 & 4.9251  & 0.9145 & 5.6594 & 121 & 11.38 & \small TIP21 \\
		& CDD-GAN \cite{chen2022unpaired}  & 12.16 & 0.25 & 0.3174 & 5.3888 & 0.8147 & 3.9248  & 0.7792 & 4.9035 & 78 & 29.27  & \small ECCV22 \\
		& D$^{4}$ \cite{yang2022self}  & 14.28 & 0.76  & 0.6903  & 4.9969 & 1.2064 & 6.7865 & 1.2227 & 7.9228 & 51 &{\bf 10.70}  &\small CVPR22 \\
		\Xhline{0.7pt} 
		\multirow{3}{*}{Paired} 
		& DAD \cite{shao2020domain} & 16.12 & 0.53 & 1.1008 & 5.5886 & 1.1837 & 5.7004 & 1.2937 & 7.0237  &- & 54.59  & \small CVPR20 \\
		& PSD \cite{chen2021psd} & 12.86 & 0.44 & 1.1304 & 4.4485 & 1.4651 & 4.5386  & 1.1578 & 5.7828 & 82 & 33.11  & \small CVPR21 \\
		& RIDCP \cite{wu2023ridcp} & 19.80   & 0.65  & 0.3105   & 3.9442      & 0.8073      & 4.5624    & 0.9440  & 4.3042      & 168  & 28.72      & \small CVPR23 \\		
		\Xhline{0.7pt} 
		\rowcolor{gray!10}Non-aligned &{\bf NSDNet} (Ours)  & {\bf 19.92} & {\bf 0.78} & {\bf 0.3031}  & {\bf 3.7686}  & {\bf 0.7621} & {\bf 3.7884} & {\bf 0.7419} & {\bf 3.6905} &{\bf  564} & 11.38 & - \\
		\Xhline{1.3pt} 
	\end{tabular}}
	\caption{Quantitative results on three real-world smoke/hazy datasets. $\downarrow$ denotes the lower the better. $\uparrow$ denotes the higher the better. Note that we only selected the latest dehazing methods (\ie, RefineNet, CDD-GAN, D$^{4}$ and PSD) for the user study.}
	\label{tab:quantitativereults}%
\end{table*}
%%%%%%%%%%%%%%%%%%%%%%%%%%%%%%%%%%%%%%%%%%%%%%%%%%%%%%%%%%%%%%%%%%%%%%%%%%%%%%

\subsection{Training Loss}
Finally, the training loss function is described as follow:
\begin{equation}
	\mathcal{L}_{\text{all}}= \mathcal{L}_{\text{msr}}+\mathcal{L}_{\text{rec}},
	\label{eq.14}
\end{equation}
where $\mathcal{L}_{\text{msr}}$ is the multi-scale reference loss in Eq.~\eqref{eq:msr}, and $\mathcal{L}_{\text{rec}}$ is the reconstruction loss. Based on the atmosphere scattering model in \eqref{eq:scatter}, $\mathcal{L}_{\text{rec}}$ is defined as:
\begin{align}
\mathcal{L}_{\text{rec}} \left(I_{\text{rec}}, I\right) =& \theta \mathcal{L}_{\ell_1}\left(I_{\text{rec}}, I\right) + \gamma \mathcal{L}_{p}\left(I_{\text{rec}}, I\right) \nonumber \\ 
&+ \eta \mathcal{L}_{\text{ssim}}\left(I_{\text{rec}}, I\right),
\label{eq:rec}
\end{align}
where $\mathcal{L}_{\ell_1}$ is the mean absolute difference loss, $\mathcal{L}_{p}$ denotes the perceptual loss~\cite{johnson2016perceptual}, $\mathcal{L}_{\text{ssim}}$ represents the structural similarity (SSIM) loss~\cite{wang2004image}. $\theta$, $\gamma$ and $\eta$ are the weight coefficients of the corresponding items, respectively. In addition, the reconstruction loss can not only supervise the training of infinite airlight, transmission, and dehazing networks but also keep the features of dehazing results independent of the non-aligned reference image. \textit{Note that the curves of the losses $\mathcal{L}_{\text{msr}}$ and $\mathcal{L}_{\text{rec}}$ are provided in supplementary material.}

\begin{figure*}[t]
	\centering
	\includegraphics[width=0.96\linewidth]{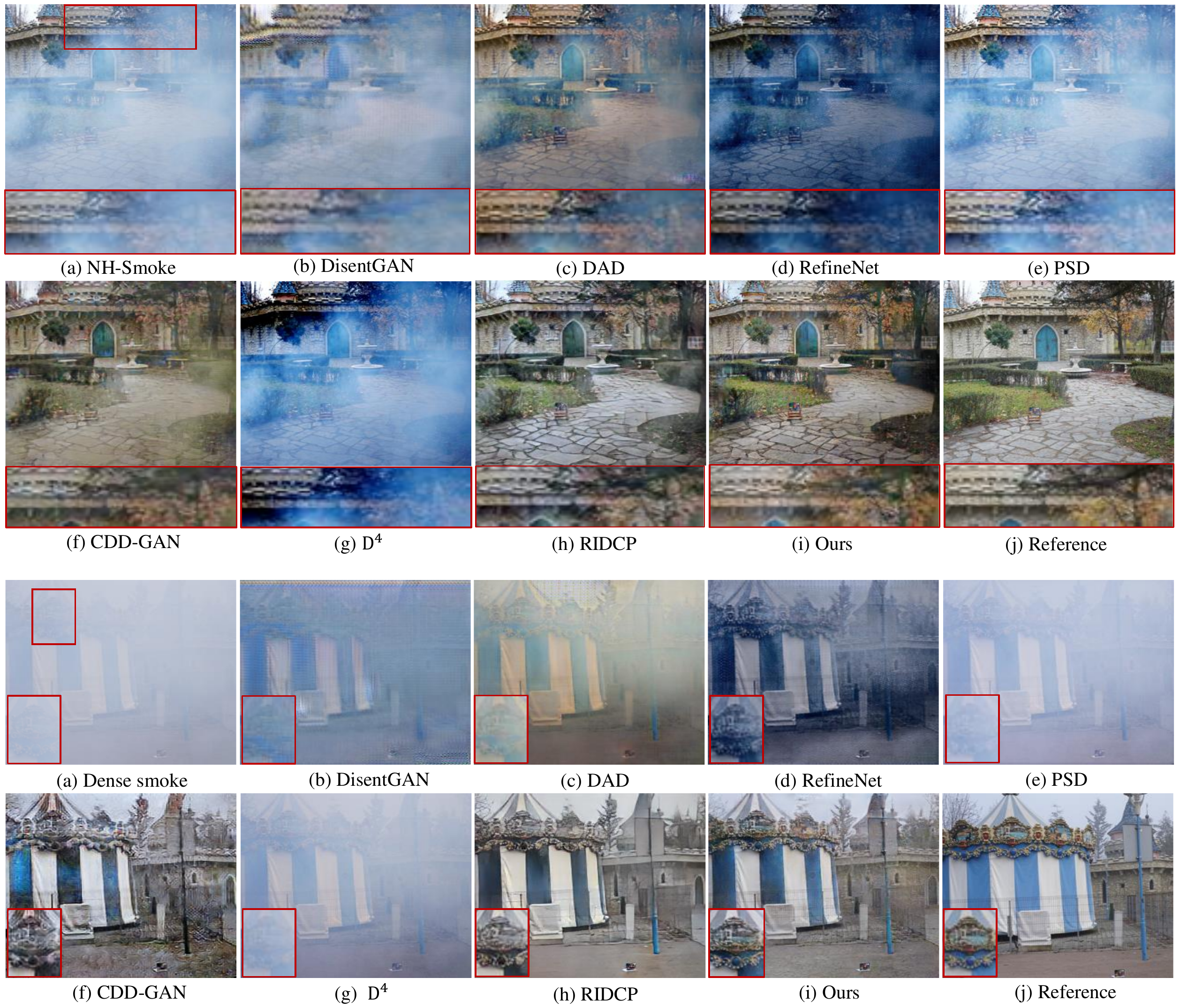}
	\caption{Dehazing results on the real-world smoke dataset. Our method effectively eliminates smoke and produces images that closely resemble the non-aligned reference. The red box indicates a zoomed-in patch, allowing for a more precise comparison.}
	\label{fig:3}
\end{figure*}

\begin{figure*}[t]
	\centering
	\includegraphics[width=0.96\linewidth]{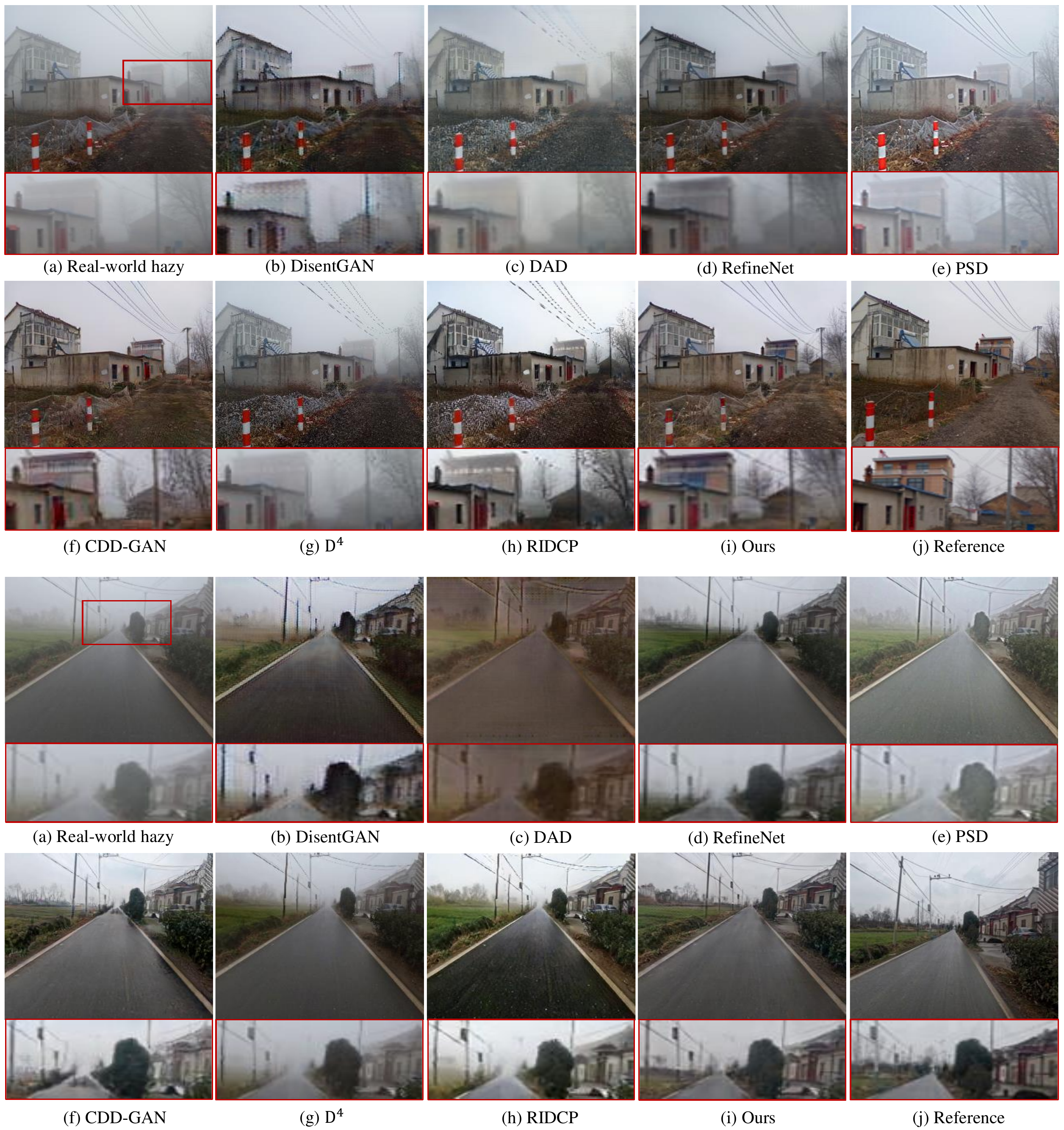}
	\caption{Dehazing results on the real-world Phone-Hazy dataset. Our method is capable of eliminating haze and producing images that closely resemble the reference image, even if they are not perfectly aligned. } %The red box corresponds to the zoomed-in patch for better comparison.
	\label{fig:4}
\end{figure*}

\begin{figure*}[t]
	\centering
	\includegraphics[width=0.95\linewidth]{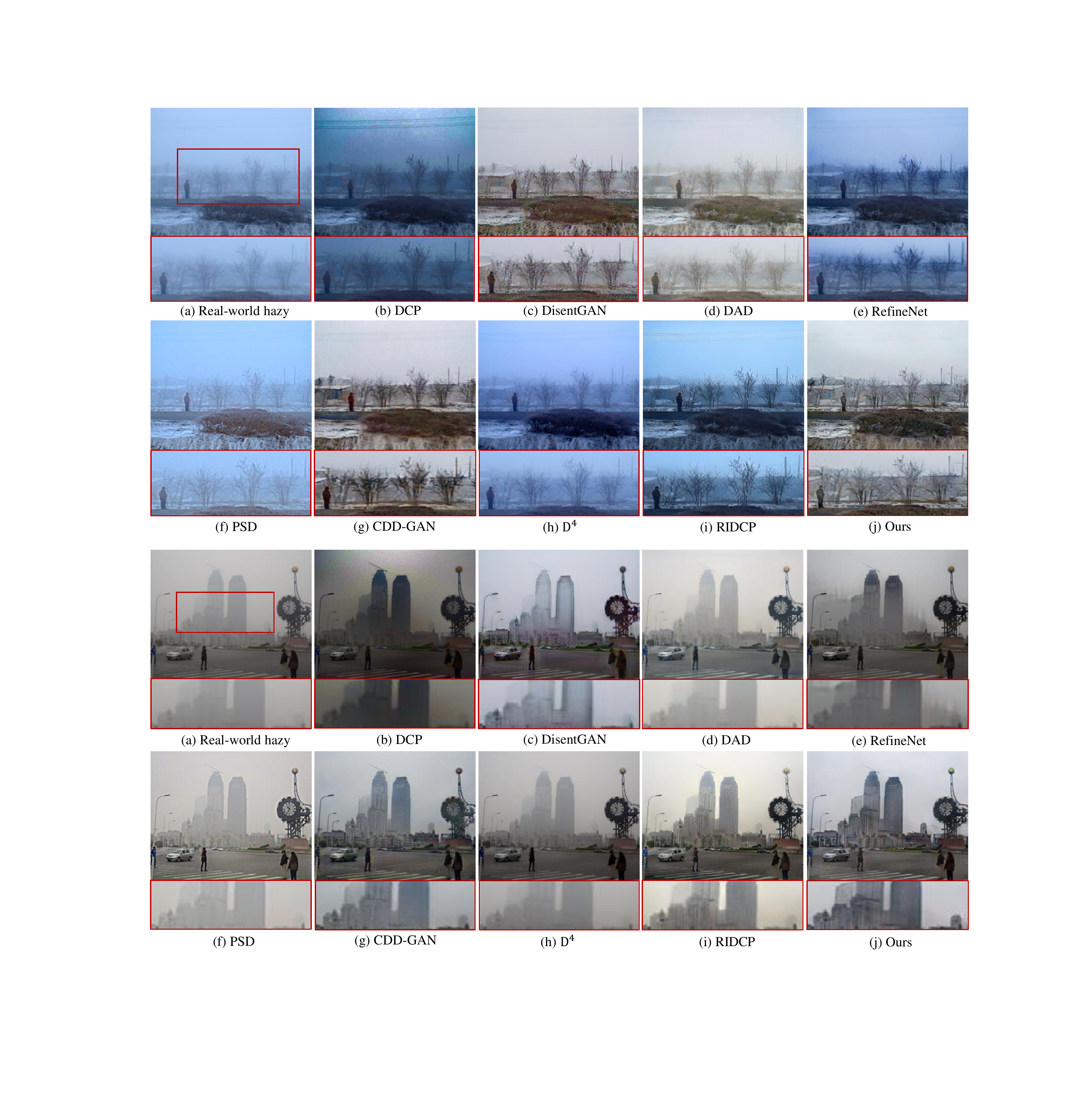}
	\caption{Dehazing results on the real-world RTTS dataset. Our method effectively eliminates haze in distant scenes while also enhancing the restoration of finer details.} %Best view by zooming.
	\label{fig:5}
\end{figure*}

\section{Experiments}

We verify the effectiveness of the proposed method by conducting experiments on three real-world smoke/hazy datasets. To further establish the effectiveness of the core modules within our proposed approach, we perform an ablation study. \emph{Note that experimental results on synthetic hazy datasets, please refer to the Appendix \ref{Appendix_B}.}
% \emph{Experimental results on synthetic dataset are \junli{provided} in the subsection \ref{Experiments on Synthetic dataset} in supplementary material}.

%\emph{Note that experimental results on synthetic hazy datasets and additional visual results of our model, please refer to the Appendix \ref{Appendix_B} and \ref{Appendix_C}.}

\subsection{Experimental Settings}

{\bf Real-world smoke Datasets}. \emph{The Real-world smoke dataset} is collected from the 2018-2021 CVPR workshops dehazing challenge. This dataset encompasses three sub-datasets: I-HAZE, O-HAZE, and NH-HAZE. It contains 155 pairs of smoke and clear images of real indoor and outdoor scenes. The smoky images encompass both homogeneous and non-homogeneous types of smoke (NH-Smoke). Besides, we resize the image of the datasets to $286\times286$, and then randomly crop images to $256\times256$. This procedure intentionally introduces misalignment between the data pairs. We randomly select 147 images for training and the remaining 8 images for the test, where the training images and test images do not overlap.

%Since our method is a non-aligned supervision, it reduces the difficulty of  is easy to  Due to the manual shooting, it is difficult to achieve pixel alignment, so the obtained data is unaligned.
\noindent \emph{Two real-world hazy datasets.} The first one, named \emph{Phone-Hazy}, involves the utilization of mobile phones (such as iPhone XR) to capture non-aligned image pairs under varying hazy and clear weather conditions at identical locations. This dataset aims to create a comprehensive real-world hazy image collection. To enhance the diversity of hazy scenes, we amassed dense hazy images from both rural and urban settings. There are a total of 415 pairs of hazy and clear images of outdoor scenes, of which 375 images were used for training and the remaining 40 images for the test. Throughout the training phase, image patches of dimensions $256\times256$ pixels were employed. Additional information regarding the collection specifics of the Phone-Hazy dataset is provided in Appendix \ref{Appendix_A:phone-hazydataset}.
The second dataset, referred to as the \emph{RTTS dataset}, is part of the RESIDE dataset\footnote{\url{https://sites.google.com/view/reside-dehaze-datasets/}} \cite{li2018benchmarking}. In this work, RTTS is employed as a third-party benchmark for evaluating dehazing methods, which contains 4322 real-world hazy images. Moreover, RTTS does not have corresponding ground truth, Consequently, utilizing this dataset to assess the performance of dehazing models ensures a fair comparison.

\noindent {\bf Implementation details}.
Firstly, in the reconstruction loss, the corresponding weight parameters $\theta$, $\beta$, and $\eta$ of $\ell_1$ loss, perceptual loss, and SSIM loss are set to 5, 1, and 1, respectively. The default value for $\omega_{1}$ and $\omega_{2}$ in Eq.~(\ref{eq:msr}) are set to 1. Secondly, in Eq.~(\ref{eq.5}), we set the relative mean $\alpha$ and relative variance $\beta$ to be 1.2 and $0.25\times10^{-3}$. In training processing, we use ADAM~\cite{kingma2015adam} optimizer with an initial learning rate of $2\times10^{-4}$. Our model was trained for 400 epochs by Pytorch with two NVIDIA GeForce RTX 3090 GPUs.

\noindent {\bf Evaluations}. In this work, we use  Fog Aware Density Evaluator (FADE)~\cite{choi2015referenceless}, Natural Image Quality Evaluator (NIQE)~\cite{mittal2012making} to evaluate the dehazing results without the ground truths (GT). In addition, we also employ PNSR \cite{huynh2008scope} and SSIM~\cite{wang2004image} to evaluate the dehazing results with the GT. %FADE is to estimate haze density of images, and take the hazy density as the evaluation of dehazing result. 

\subsection{Results on real smoke/hazy datasets}

To evaluate the generalization and effectiveness of our NSDNet on real smoke/hazy scenes, we compare with some state-of-the-art (SOTA) methods including DCP~\cite{he2010single}, DisentGAN~\cite{yang2018towards}, DAD~\cite{shao2020domain}, RefineNet~\cite{zhao2021refinednet}, PSD~\cite{chen2021psd}, CDD-GAN~\cite{chen2022unpaired}, RIDCP~\cite{wu2023ridcp} and D$^{4}$~\cite{yang2022self}. For fair comparisons, we fine-tune each method on the real smoke/hazy datasets to achieve their best performances. Table~\ref{tab:quantitativereults} summarizes the results of the quantitative comparison.

\emph{Results on the real-world smoke dataset.} From the Table~\ref{tab:quantitativereults}, we found that our NSDNet outperforms all state-of-the-art (SOTA) methods in quantitative metrics..
For instance, when compared to the unpaired DCP method \cite{he2010single}, our approach demonstrates significant improvements, with a PSNR increase of 4.48, SSIM enhancement of 0.39, FADE improvement of 0.093, and a NIQE reduction of 0.3617. In contrast, in comparison to the paired RIDCP method~\cite{wu2023ridcp}, our approach still achieves notable improvements, with a PSNR gain of 0.12, SSIM increase of 0.13, FADE enhancement of 0.0074, and a NIQE reduction of 0.17. 

Furthermore, as illustrated in Fig.~\ref{fig:3}, the visual restoration results for the smoke image are presented. When compared to state-of-the-art methods, it becomes apparent that our NSDNet exhibits a closer resemblance to the clear reference image in terms of both color and texture. These methods, often designed with synthetic datasets in mind and lacking the constraints of physical priors, typically struggle to effectively remove smoke. For instance, RefineNet~\cite{zhao2021refinednet} produces a darker dehazing result and retains a small amount of smoke. Similarly, the DisentGAN~\cite{yang2018towards}, DAD~\cite{shao2020domain}, PSD~\cite{chen2021psd}, and D$^{4}$~\cite{yang2022self} methods do not effectively remove heavy smoke. Additionally, CDD-GAN~\cite{chen2022unpaired} generates blurrier textures and experiences color distortion, while RIDCP \cite{wu2023ridcp} yields excessively smooth dehazing results that also suffer from color distortion, such as the grey tent.

%\emph{More visual results are shown in Figs.\ref{fig:S3}-\ref{fig:S4} of supplementary material.}

%\subsection{Experiments on real-world Phone-Hazy dataset}
\emph{Results on the real-world Phone-Hazy dataset collected by us.} Fig.~\ref{fig:4} showcases the visualizations of our dehazing results, highlighting the superior performance of our NSDNet compared to state-of-the-art (SOTA) methods in terms of both brightness and texture details. Specifically, when compared to the DisentGAN \cite{yang2018towards}, DAD~\cite{shao2020domain}, RefineNet~\cite{zhao2021refinednet}, and D$^{4}$ \cite{yang2022self} methods, our NSDNet not only eliminates persistent heavy hazes but also mitigates artifacts. While PSD~\cite{chen2021psd} enhances brightness, it falls short of completely eliminating haze. On the other hand, CDD-GAN \cite{chen2022unpaired} and RIDCP~\cite{wu2023ridcp} exhibit promising haze removal capabilities near the camera but struggle with texture and color quality in their dehazed results. Moreover, they fail to restore scenes distant from the camera, as exemplified by the branches and the road in Fig.~\ref{fig:4}. Additionally, RIDCP~\cite{wu2023ridcp} stands out for its overly bright colors. In summary, NSDNet excels in restoring finer details and producing visually appealing images, particularly in terms of scene brightness restoration. 

Given the absence of aligned ground truths, we employ the NIQE and FADE metrics for evaluation in Table~\ref{tab:quantitativereults}. These metrics underscore the exceptional performance of our NSDNet, as it achieves the lowest NIQE and FADE values. For instance, our approach surpasses the unpaired CDD-GAN method~\cite{chen2022unpaired}  with improvements of 0.0526 in FAQE and 0.1364 in NIQE. When compared to the paired RIDCP method~\cite{wu2023ridcp} , NSDNet exhibits even more significant enhancements, with 0.0452 in FAQE and 0.7740 in NIQE.

% Specifically, DisentGAN \cite{yang2018towards}, DAD \cite{shao2020domain}, RefineNet~\cite{zhao2021refinednet}, and D$^{4}$ \cite{yang2022self} not only remain the lingering heavy hazes, but also exhibit some artifacts. PSD \cite{chen2021psd} enhances brightness, but it falls short of eliminating the haze completely. Although CDD-GAN \cite{chen2022unpaired} and RIDCP \cite{wu2023ridcp} demonstrate promising haze removal capabilities, the texture and color quality of the dehazed results remain less than ideal, with RIDCP \cite{wu2023ridcp} exhibiting overly bright colors. Overall, our NSDNet excels in restoring finer details and delivering visually pleasing images, particularly when it comes to restoring scene brightness. Due to no aligned ground truths, in addition, we use the NIQE and FADE metrics to evaluate the methods, and Table~\ref{tab:quantitativereults} shows the results. We can see that our NSDNet obtains the lowest NIQE and FADE values, demonstrating its excellent performance. For example, our approach is better than the unpaired CDD-GAN \cite{chen2022unpaired} method with the 0.0526 FAQE and 0.1364 NIQE improvements. Compared to the paired RIDCP \cite{wu2023ridcp} method, our NSDNet also has 0.0452 FAQE and 0.7740 NIQE improvements. 

%\emph{More visual results are shown in Figs. \ref{fig:S5}-\ref{fig:S6} of supplementary material}.

\begin{figure*}[t]
	\centering
	\includegraphics[width=0.96\linewidth]{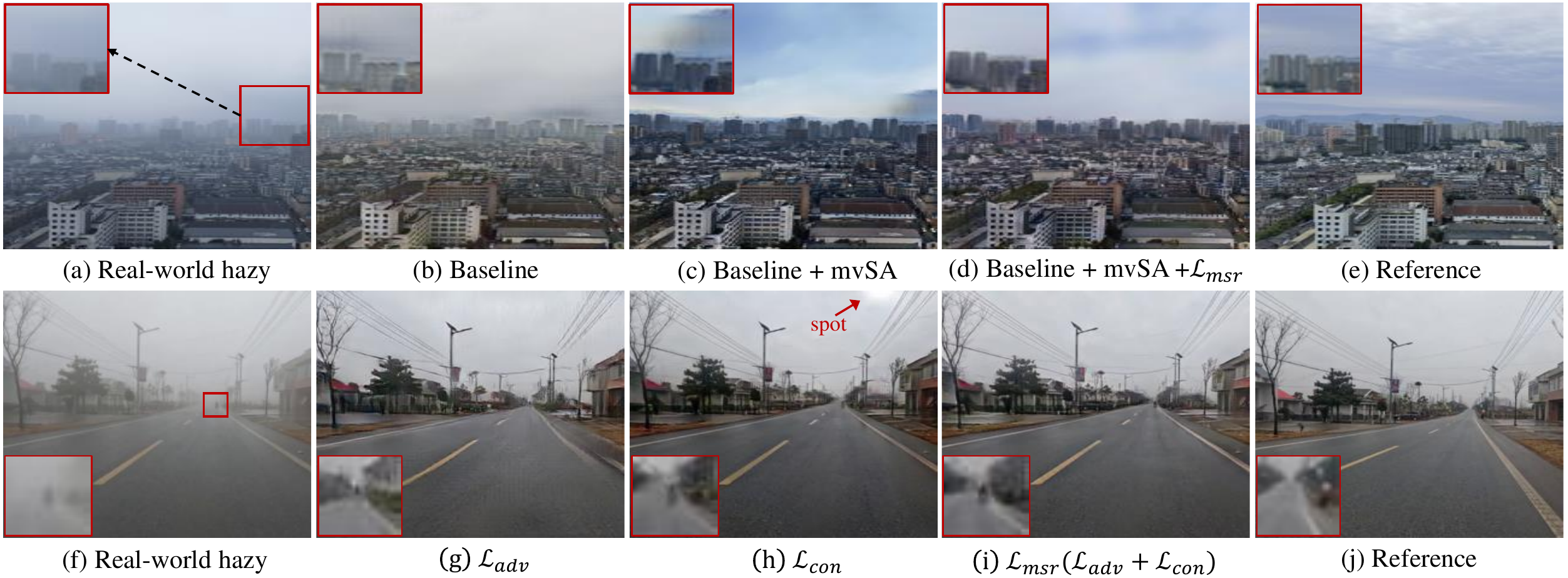}
	\caption{In the top line, (a)-(e) show the effectiveness of mvSA and $\mathcal{L}_{\text{msr}}$ on a real dense hazy image with long-range view (\emph{challenging scene}). In the bottom line, (f)-(j) showcase the visualization of the ablation study for $\mathcal{L}_{\text{msr}}$.}
	\label{fig:6}
\end{figure*}

\emph{Results on the real-world RTTS dataset.} To further assess the generalization prowess of our NSDNet, all methods were exclusively evaluated on the RTTS dataset, and their quantitative comparisons are detailed in Table~\ref{tab:quantitativereults}. It is evident that NSDNet outperforms its counterparts, exhibiting the lowest NIQE and FADE scores. Specifically, our approach outperforms the unpaired CDD-GAN method~\cite{chen2022unpaired} by showing improvements of 0.0373 in FADE and 1.2130 in NIQE. Furthermore, our approach outpaces the paired RIDCP method \cite{wu2023ridcp} by achieving significant improvements of 0.2021 in FADE and 0.6137 in NIQE. This can be attributed to the fact that RIDCP utilizes predictive depths to synthesize hazy and clear image pairs for model training. However, these predictive depths tend to be more accurate in close scenes and less reliable in distant ones. Consequently, the trained models excel at haze removal in close scenes but struggle when dealing with far scenes, resulting in elevated FADE and NIQE scores. Furthermore, our NSDNet demonstrates its exceptional image dehazing capabilities through a significantly superior user study.

Additionally, Fig. \ref{fig:5} provides a visual representation of our findings. The following observations can be made: 1) RefineNet~\cite{zhao2021refinednet}, PSD~\cite{chen2021psd}, D$^{4}$ \cite{yang2022self}, and RIDCP~\cite{wu2023ridcp} struggle to effectively remove haze in distant objects, such as the door and the building. 2) Although CDD-GAN \cite{chen2022unpaired} visually resembles our NSDNet, it falls short in restoring images with satisfactory texture and color. 3) In contrast, our NSDNet not only eliminates haze in distant objects but also restores brightness and texture details with exceptional proficiency.

\emph{Analysis influencing subpar performances.} To shed light on the reasons behind suboptimal performance analysis, we can pinpoint several critical issues. Firstly, the widely recognized DCP~\cite{he2010single} faces challenges in effectively handling sky regions and exhibits sensitivity to crucial parameters, such as constraints on $A_{\infty}$ and $t$ boundaries. Secondly, unpaired unsupervised learning approaches, exemplified by methods like DisentGAN~\cite{yang2018towards}, RefineNet~\cite{zhao2021refinednet}, CDD-GAN~\cite{chen2022unpaired}, and D$^{4}$ \cite{yang2022self}, employ GANs to generate dehazed images. These GANs are trained using unpaired data, which, unfortunately, are drawn from different haze and clear image distributions. This domain inconsistency poses challenges during model training, ultimately resulting in suboptimal performance. Thirdly, paired methods, such as DAD~\cite{shao2020domain}, PSD~\cite{chen2021psd}, and RIDCP~\cite{wu2023ridcp}, often incorporate domain adaptation techniques to train dehazing models on synthetically aligned data while testing on real hazy images. This practice introduces a domain gap between synthetic and real data. Additionally, PSD~\cite{chen2021psd} employs pseudo-labels generated by Contrast Limited Adaptive Histogram Equalization (CLAHE) for fine-tuning, which can lead to excessively vibrant colors in the dehazed output.

\textit{Note that additional visual image and video dehazing results can be found in the supplementary materials.}

%Firstly, as is well known, DCP \cite{he2010single} struggles to handle sky regions and is sensitive to critical parameters (\ie, constraints on $A_{\infty}$ and $t$ boundaries). Secondly, the unpaired unsupervised learning methods, likes DisentGAN \cite{yang2018towards}, RefineNet \cite{zhao2021refinednet}, CDD-GAN \cite{chen2022unpaired}, and D$^{4}$ \cite{yang2022self} utilize GANs to generate dehazed images. The GANs are trained by using unpaired data. However, these unpaired data are sampled from the different haze and clear image distributions. This domain inconsistency makes the models difficultly training, leading to suboptimal performance. Thirdly, the paired methods, likes DAD \cite{shao2020domain}, PSD \cite{chen2021psd} and RIDCP \cite{wu2023ridcp} often employ domain adaptation technologies to train the dehazed models on synthetic alignment data, and to test on the real haze images, resulting in a domain gap between the synthetic and real data. In addition, PSD \cite{chen2021psd} utilizes pseudo-labels generated by CLAHE for fine-tuning the model, leading to excessively vibrant colors in the dehazed results.

\begin{table}[t]
	\linespread{1.0}
	\centering
	\setlength\tabcolsep{1.8pt} 
	\renewcommand\arraystretch{1.4}
	\scalebox{0.90}{	
		\begin{tabular}{l|cc|cccc}
			\Xhline{1.3pt} 
			\multicolumn{1}{c|}{\multirow{2}[1]{*}{\makecell{Ablations for\\core module}}} & \multicolumn{2}{c|}{Phone-Hazy} & \multicolumn{4}{c}{Real-world Smoke} \\
			%\cline{2-7} 
			& \multicolumn{1}{c}{FADE$\downarrow$} & \multicolumn{1}{c|}{NIQE$\downarrow$} & \multicolumn{1}{c}{FADE$\downarrow$} & \multicolumn{1}{c}{NIQE$\downarrow$}& \multicolumn{1}{c}{PSNR$\uparrow$} & \multicolumn{1}{c}{SSIM$\uparrow$}\\
			\Xhline{0.7pt} 
			Baseline      									   & 0.8623   & 5.1218   & 0.3512  & 3.9455  & 16.48 & 0.55 \\
			%+SA												   &		  &			 &         &         &       &      \\
			+mvSA 										       & 0.8387   & 4.7817   & 0.3403  & 3.9087  & 17.01 & 0.57 \\
			\rowcolor{gray!10}+mvSA+$\mathcal{L}_{\text{msr}}$ & \textbf{0.7621} & \textbf{3.7884} & \textbf{0.3031} & \textbf{3.7686} & \textbf{19.92} & \textbf{0.78}\\  
			\Xhline{1.3pt} 
	\end{tabular}}
	\caption{Quantitative results of ablation study on real-world smoke and Phone-Hazy dataset.}
	\label{table2:ablation study}%
\end{table}%

\subsection{Ablation Study}
{\bf Effect of mvSA and $\mathcal{L}_{\text{msr}}$.} To assess the effectiveness of the mvSA network and the multi-scale reference loss $\mathcal{L}_{\text{msr}}$, we conducted a series of ablation experiments to evaluate our approach on real-world smoke and Phone-Hazy datasets. We constructed a dehazing framework \emph{baseline}, which consists of two deep networks for processing clear scenes ($J$) and estimating the transmission map ($t$), along with a deep network that generates a constant infinite airlight using a U-Net architecture. This baseline was trained using both reconstruction loss and adversarial loss. Subsequently, we replaced the DCP method with the mvSA network and introduced $\mathcal{L}_{\text{msr}}$ to train the dehazing network, resulting in two variations: \emph{baseline+mvSA} and \emph{baseline+mvSA+$\mathcal{L}_{\text{msr}}$} (our NSDNet). The quantitative results can be found in Table~\ref{table2:ablation study}. Notably, \emph{baseline+mvSA+$\mathcal{L}_{\text{msr}}$} achieved the lowest FADE and NIQE values, the highest PSNR and SSIM values, underscoring the exceptional performance of our NSDNet in real image dehazing.

Furthermore, in Fig.~\ref{fig:6}(a)-(e), we present dehazing visualizations of a challenging hazy image with an expansive field of view. The \emph{baseline} approach exhibits color distortion and retains numerous haze residuals due to its inaccurate estimation of the infinite airlight ($A_{\infty}$). Conversely, the inclusion of the \emph{baseline+mvSA} method results in improved scene recovery, as demonstrated in (c), primarily because \emph{mvSA} effectively estimates variations in the infinite airlight. Most notably, (d) showcases that \emph{baseline+mvSA+$\mathcal{L}_{\text{msr}}$} produces noticeably clearer and more aesthetically pleasing dehazing outcomes, such as enhanced texture in the sky area and distant buildings, when compared to \emph{baseline+mvSA}. This serves as strong evidence supporting the effectiveness of the proposed $\mathcal{L}_{\text{msr}}$ technique.

\begin{figure}[t]
	\centering
	\includegraphics[width=0.96\linewidth]{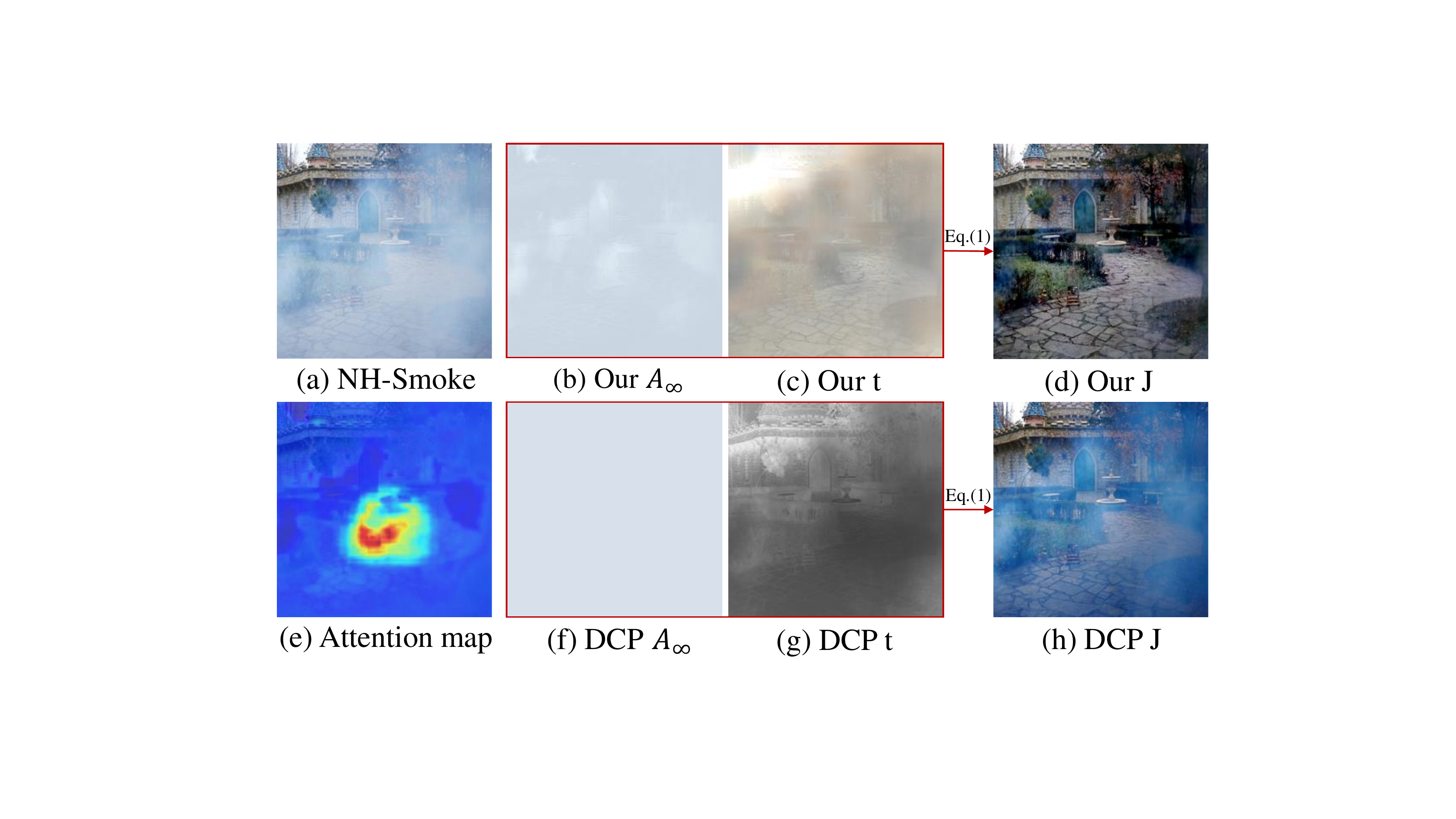}
	\caption{Visualizations of the dehazing result $J$, the infinite airlight $A_{\infty}$, the attention map and the transmission map $t$. }
	\label{fig:7}
\end{figure}

To further validate the capability of the \emph{mvSA} model in learning more accurate infinite airlight values, we conducted a comparative visualization using a real smoke image in Fig.~\ref{fig:7} 
comparing it with the widely-used DCP method, which assumes a constant infinite airlight. Fig.~\ref{fig:7}(a) shows a non-homogeneous smoke (NH-Smoke) scene, while Fig.~\ref{fig:7}(b) showcases the $A_{\infty}$ prediction generated by \emph{mvSA}. This prediction exhibits a more accurate representation of the realistic variations in comparison to the DCP method shown in Fig.~\ref{fig:7}(f). Furthermore, in Fig.~\ref{fig:7}(d) and Fig.~\ref{fig:7}(h), we compare the dehazing results achieved by our \emph{mvSA} network in terms of $A_{\infty}$ and transmission ($t$) using Eq.~\eqref{eq:scatter} against those obtained using the DCP method. Our approach clearly demonstrates significantly improved dehazing results. Additionally, we provide a visualization of the attention map in Fig.~\ref{fig:7}(e), which is based on the dark channel prior and the hazy image. This attention map highlights regions corresponding to heavy haze, showcasing the effectiveness of the \emph{mvSA} network in capturing and addressing challenging hazy conditions.

\begin{table}[t]
	\linespread{1.0}
	\centering
	\setlength\tabcolsep{1.4pt}%???
	\renewcommand\arraystretch{1.4}
	\scalebox{1.0}{
		\begin{tabular}{l|c|cc|cc}
			\Xhline{1.3pt} 
			Misalignment & Scale & \multicolumn{1}{c}{PSNR $\uparrow$} & \multicolumn{1}{c|}{SSIM $\uparrow$} &
			\multicolumn{1}{c}{FADE $\downarrow$} & \multicolumn{1}{c}{NIQE $\downarrow$}\\
			\Xhline{0.7pt} 
			0 pixels   & 0         & \textbf{20.14} & \textbf{0.78}  & \textbf{0.2942}  & \textbf{3.4936}   \\
			\rowcolor{gray!20} 30 pixels (Ours) & 0.117   & 19.92 		 & 0.78           & 0.3031  & 3.7686   \\
			60 pixels   &    0.234    & 19.62          & 0.76           & 0.3225  & 3.8819   \\
			90 pixels  &   0.351      & 19.42          & 0.76           & 0.3107  & 3.9891   \\
			120 pixels   &  0.468     & 18.45          & 0.73           & 0.3442  & 4.1253   \\
			\Xhline{1.3pt} 
	\end{tabular}}
	\caption{Comparison of various misalignment pixel scenarios using the real-world smoke dataset with dimensions of 256x256 pixels.}
	\label{tab3:non-aligned scale}
\end{table}

\textbf{Enhanced scale selection for misalignment and rotation.}
In this section, our primary focus lies on assessing the impact of misalignment and rotation in the context of non-aligned reference images from a real-world smoke dataset. \textit{Investigating misalignment effects:} To comprehensively evaluate the effects of misalignment, we introduced varying misalignment levels ranging from 0 to 120 pixels. The resulting performance metrics, including PSNR, SSIM, FADE, and NIQE, are presented in Table~\ref{tab3:non-aligned scale}. These results consistently highlight the direct relationship between reduced misalignment and improved model performance. Notably, our experiments were conducted with a 30-pixel misalignment on a smoke dataset, while the Phone-Hazy dataset exhibited more significant misalignment issues, inconsistent focal lengths, and varying perspectives.

\begin{figure}[t]
	\centering
	\includegraphics[width=0.96\linewidth]{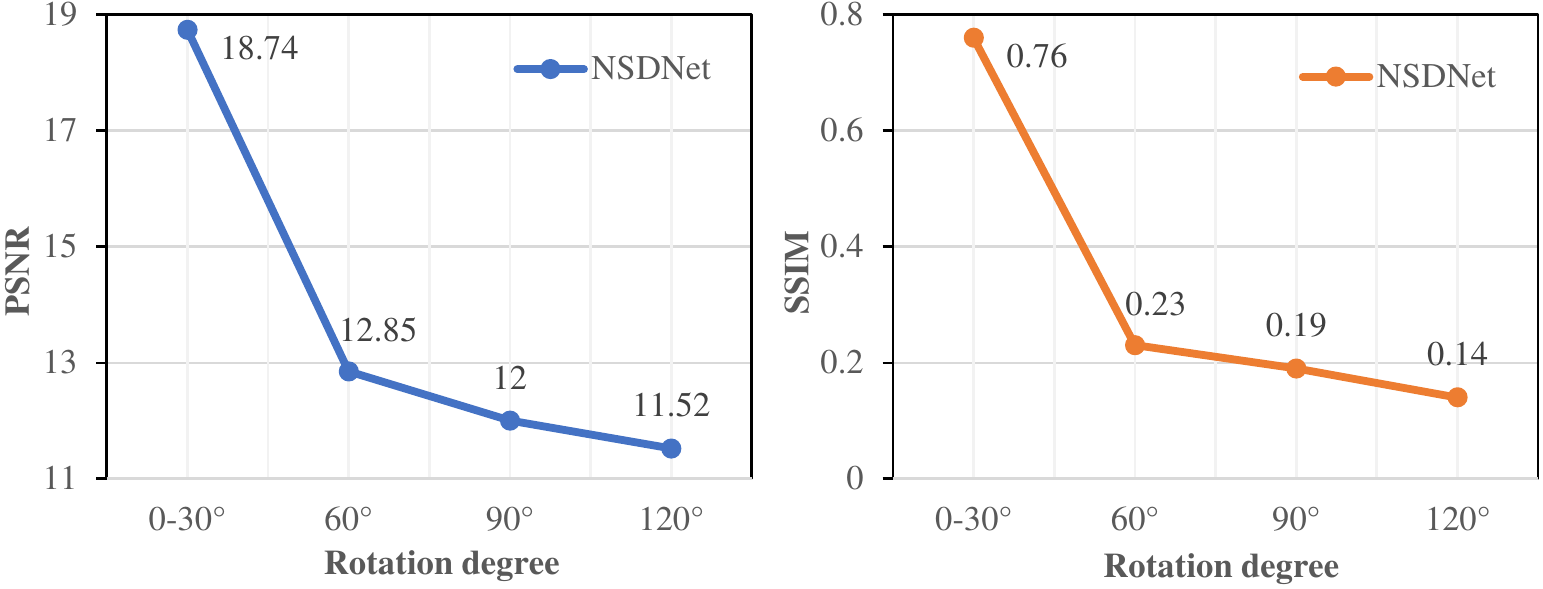}
	\caption{PSNR and SSIM results across various rotation angles using the real-world smoke dataset.}
	\label{fig:10}
\end{figure}

\begin{table}[t]
	\linespread{1.0}
	\centering
	\setlength\tabcolsep{2.1pt}%???
	\renewcommand\arraystretch{1.4}
	\scalebox{1.0}{
		\begin{tabular}{c|cc|cc}
			\Xhline{1.3pt}
			\multicolumn{5}{c}{\makecell{Fix $\mathcal{L}_{\ell_1}:\mathcal{L}_{p}:\mathcal{L}_{ssim}$ to $5:1:1$}}  \\
			\Xhline{0.7pt}
			$\mathcal{L}_{msr}$  & \multicolumn{1}{c}{PSNR $\uparrow$} & \multicolumn{1}{c|}{SSIM $\uparrow$} & \multicolumn{1}{c}{FADE $\downarrow$} & \multicolumn{1}{c}{NIQE $\downarrow$}\\
			\Xhline{0.7pt}
			$\mathcal{L}_{\text{msa}}$             & 17.08       & 0.58          & 0.3328 & 3.8936\\
			$\mathcal{L}_{\text{msc}}$             & 18.04       & 0.76          & 0.3146 & 3.8037\\
			\rowcolor{gray!10}$\mathcal{L}_{\text{msa+msc}}$ (Ours)  & {\bf 19.92} & {\bf 0.78}    & \textbf{0.3031} & \textbf{3.7686}\\
			\Xhline{0.7pt}
			\Xhline{0.7pt}
			\multicolumn{5}{c}{\makecell{Fix $\mathcal{L}_{msa}:\mathcal{L}_{msc}$ to $1:1$}}  \\
			\hline
			$\mathcal{L}_{rec}$ & \multicolumn{1}{c}{PSNR $\uparrow$} & \multicolumn{1}{c|}{SSIM $\uparrow$} & \multicolumn{1}{c}{FADE $\downarrow$} & \multicolumn{1}{c}{NIQE $\downarrow$}\\
			\Xhline{0.7pt}
			$\mathcal{L}_{\ell_1} (base)$             & 19.11          & 0.76           & 0.3173 & 3.8168 \\
			$\mathcal{L}_{\ell_1 + p}$                & 19.36          & 0.77           & 0.3158 & 3.7710 \\
			\rowcolor{gray!10}$\mathcal{L}_{\ell_1 + p + ssim}$ (Ours)  & \textbf{19.92} & \textbf{0.78}  & \textbf{0.3031} & \textbf{3.7686} \\
			\Xhline{1.3pt}
	\end{tabular}}%
	\caption{An ablation study was conducted on a real-world smoke dataset to investigate the effects of different loss components, specifically focusing on $\mathcal{L}{\text{msr}}$ and $\mathcal{L}{\text{rec}}$.}
	\label{tab5:different loss items}
\end{table}%

\textit{Exploring rotation effects:} Furthermore, to examine the impact of rotation on non-aligned reference images, we employed rotation angles of 30\degree, 60\degree, and 90\degree. Fig.~\ref{fig:10} visually illustrates the outcomes of rotating non-aligned reference images in increments of 0 to 30\degree, simulating real-world scenarios. It is worth noting that the model's performance exhibits sensitivity to variations in the rotation angle, primarily due to the influence of pixel positions on contextual loss. However, in practical data acquisition, it is relatively straightforward to constrain the rotation angle of captured images within the range of 0 to 30\degree.

\begin{table}[t]
	\linespread{1.0}
	\centering
	\setlength\tabcolsep{2.1pt}%???
	\renewcommand\arraystretch{1.4}
	\scalebox{1.0}{
		\begin{tabular}{c|ccc|cc}
			\Xhline{1.3pt}
			The scale of $\mathcal{L}_{\text{msr}}$ & \multicolumn{1}{l}{0.5$\times$} & 1$\times$ & 2$\times$ & \multicolumn{1}{l}{FADE $\downarrow$}  & NIQE $\downarrow$ \\
			\Xhline{0.7pt}
			$1\times$                 &       & $\checkmark$     &                                   & 0.7921     & 4.1225 \\
			$0.5\times$ \& $1\times$          & $\checkmark$     & $\checkmark$     &                & 0.8043     & 3.9782 \\
			$1\times$ \& $2\times$            &       & $\checkmark$     & $\checkmark$              & 0.7786     & 4.3201 \\
			\rowcolor{gray!10}$0.5\times$ \& $1\times$ \& $2\times$ (Ours)   & $\checkmark$ & $\checkmark$ & $\checkmark$    & {\bf 0.7621} & {\bf 3.7884} \\
			\Xhline{1.3pt}
	\end{tabular}}
	\caption{Comparison of different scales on Phone-Hazy dataset for $\mathcal{L}_{\text{msr}}$. $\&$ denotes "and".}
	\label{tab4:Scale selection}
\end{table}

{\bf Effect of the losses $\mathcal{L}_{\text{msr}}$ and $\mathcal{L}_{\text{rec}}$.} The reference loss $\mathcal{L}_{\text{msr}}$ in the Eq.~\eqref{eq:msr} includes $\mathcal{L}_{msa}$ and $\mathcal{L}_{msc}$, and the reconstruction loss $\mathcal{L}_{\text{rec}}$ in the Eq.~\eqref{eq:rec}, 
consists of $\mathcal{L}_{1}$, $\mathcal{L}_{p}$ and $\mathcal{L}_{ssim}$. Here, we employ the distinct losses to train our model on the real-world smoke dataset. The quantitative results are shown in Table~\ref{tab5:different loss items}. In the upper section of the table, we highlight the efficacy of $\mathcal{L}_{\text{msa}}$ and $\mathcal{L}_{\text{msc}}$. In particular, $\mathcal{L}_{\text{msr}}$ achieves a better performance than $\mathcal{L}_{\text{msa}}$ and $\mathcal{L}_{\text{msc}}$ when fixed the loss $\mathcal{L}_{\text{rec}}$. Furthermore, we visualize the results of ablating $\mathcal{L}_{\text{msr}}$, as shown in Fig.~\ref{fig:6} (f) - (j). The down part of the table also showcases the similar effectiveness of $\mathcal{L}_{1}$, $\mathcal{L}_{p}$ and $\mathcal{L}_{ssim}$. In conclusion, these ablations provide evidence that these loss components are valuable for enhancing detail restoration and improving image dehazing performance in real-world scenarios.

{\bf Effect of scale selection for $\mathcal{L}_{\text{msr}}$}. The proposed $\mathcal{L}_{\text{msr}}$ loss function encompasses a range of different scales. To assess the impact of these various scales, we conducted experiments with three specific scales (0.5$\times$, 1$\times$, 2$\times$). The results displayed in Table~\ref{tab4:Scale selection} demonstrate that utilizing a multi-scale approach can effectively enhance the performance of $\mathcal{L}_{\text{msr}}$. Empirically, we opted for a three-scale setup (0.5$\times$, 1$\times$, 2$\times$) to strike a balance between performance improvement and computational complexity.

\begin{figure}[t]
	\centering
	\includegraphics[width=0.96\linewidth]{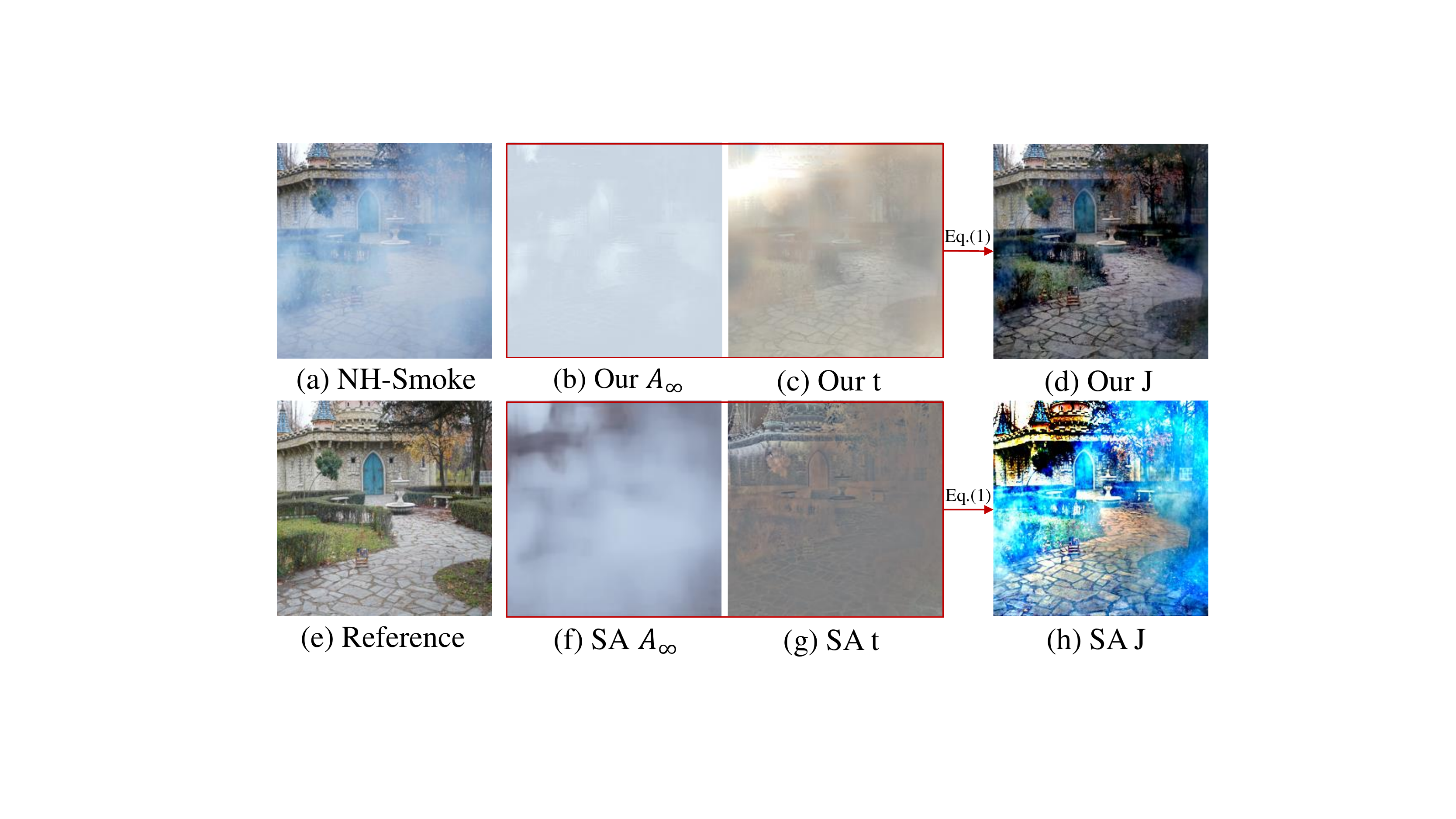}
	\caption{Visual comparison of $A_{\infty}$, t and J of mvSA and SA.}
	\label{fig:8}
\end{figure}

\section{Discussions and Analysis}

%\noindent {\bf Why our framework works ?} There are three explanations. 1) Our definitions of infinite airlight $A_{\infty}$ and transmission map $t$ are closer to the real scene. So, our $A_{\infty}$ and $t$ can generate dehazing results with better visual quality are shown in Fig.~\ref{fig:7}, thus producing a better restraint effect on dehazing result $J$. 2) We designed multi-scale reference loss to enable our framework more accurately recover the brightness and texture of haze-free scenes by referencing the non-aligned images of the corresponding scene. 3) Our framework is more reasonable and effective by constraints of physical priors (\ie, dark channel, atmospheric scattering model).

\begin{figure}[t]
	\centering
	%\vskip -0.1in
	\includegraphics[width=0.96\linewidth]{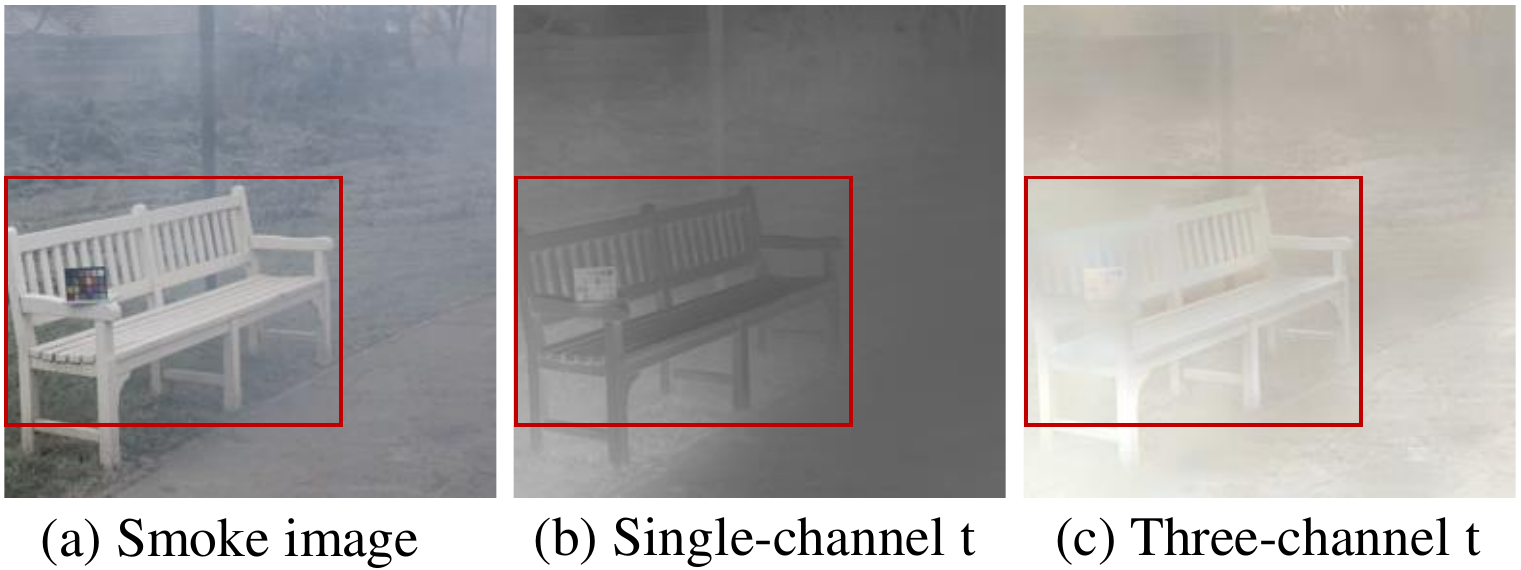}
	\caption{Comparing the transmission map $t$ in single-channel and three-channel transmission.}
	\label{fig:9}
\end{figure}

\begin{figure*}[ht]
	\centering
	\includegraphics[width=0.96\linewidth]{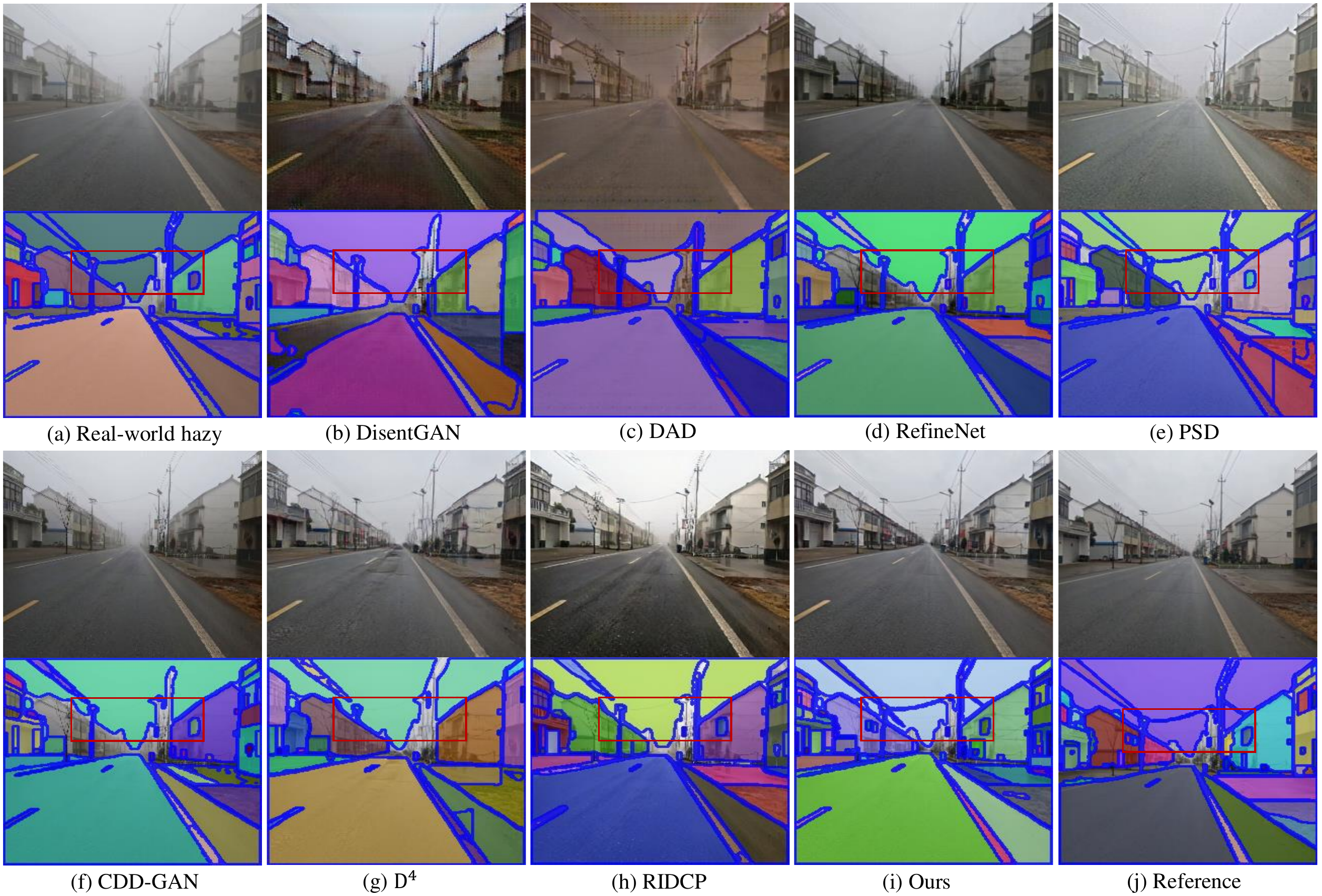}
	\caption{Visual results of semantic segmentation on the Phone-Hazy dataset. Our results excelled in window and electric wire locations.}
	\label{fig:11}
\end{figure*}

\textbf{Why do the dehazing results look blurry in the real world?} From the above dehazing visualizations, we can see that the dehazed outcomes exhibit blurriness when applied to real-world images. One plausible explanation for this phenomenon lies in the degradation of hazy images due to particle scattering during their capture by the camera \cite{nayar1999vision,zhang2023weatherstream}. However, many existing models focus solely on the dehazing task, neglecting the critical aspect of image restoration. Consequently, when compared to the reference image, the dehazing results exhibit noticeable blurriness in real-world scenarios. In contrast, synthetic data yield superior results. This can be attributed to the synthetic data being generated by introducing haze to clear images, thereby avoiding any degradation in image sharpness. To gain a clearer understanding of the dehazing results achieved with synthetic data, please refer to the visualization provided in Appendix Fig.~\ref{fig:Appendix_B1}.

\textbf{Is self-attention (SA) more effective than our modified variant of SA (mvSA) when dealing with non-uniform $A_{\infty}$ distributions?} No, it is not. The primary reason is that SA frequently generates attention maps that reveal numerous ineffectual regions within $A_{\infty}$. In contrast, mvSA utilizes the DCP strategy to identify the top 1\% of the most significant points in the attention map and computes their mean. We illustrate this difference in Fig.~\ref{fig:8}.

\textbf{Which option is more suitable for the transmission map, single-channel or three-channel?} According to the formula $t(x) = e^{-\beta(\lambda)d(x)}$, there are two scenarios where the transmission map tends to approach zero: one is in hazy areas, and the other is in the sky region at infinity (\ie, $d(x)$ approaches $\infty$). Examining Fig.~\ref{fig:9} (b), it is evident that the transmission map for the white chair marked by the red box should not approach zero in the case of the single-channel transmission map. In contrast, the three-channel transmission map is comparatively more accurate, leading to superior dehazing results.

%\begin{table}[t]
%	\centering
%	\setlength\tabcolsep{7.0pt}%???
%	\renewcommand\arraystretch{1.3}
%	\scalebox{1.0}{
%		\begin{tabular}{l|>{\columncolor{gray!10}}c|c|c|c}
%			\hline
%			Rotation& \multicolumn{1}{>{\columncolor{gray!10}}c|}{0$\sim$30\degree(Ours)} & \multicolumn{1}{c|}{30\degree} & \multicolumn{1}{c|}{60\degree}& \multicolumn{1}{c}{90\degree}\\
%			\hline
%			PSNR $\uparrow$ &{\bf18.74} &12.85 &12.00 &11.52 \\
%			SSIM $\uparrow$ &{\bf 0.76} & 0.23 &0.19 &0.14  \\
%			\hline
%	\end{tabular}}
%	\caption{Comparing reference images at different rotation degree on the real-world smoke dataset.}
%	\label{tabS5:Rotation}
%	\vskip -0.1in
%\end{table}%

\textbf{Advantages for downstream tasks.} To highlight the benefits of enhancing real-world images by reducing haze for subsequent tasks, we employed the FastSAM tool\footnote{\url{https://replicate.com/casia-iva-lab/fastsam}}~\cite{zhao2023fast}. We used it to assess the advantages of different dehazing models in the context of image segmentation. As depicted in Fig.~\ref{fig:11}, our dehazing results demonstrate the ability to segment smaller objects (such as windows and electric wires) more effectively when compared to other state-of-the-art dehazing techniques. This improved performance is attributed to our capability to mitigate haze over greater distances and restore finer texture details and scene brightness.

\begin{figure}[t]
	\centering
	\includegraphics[width=0.96\linewidth]{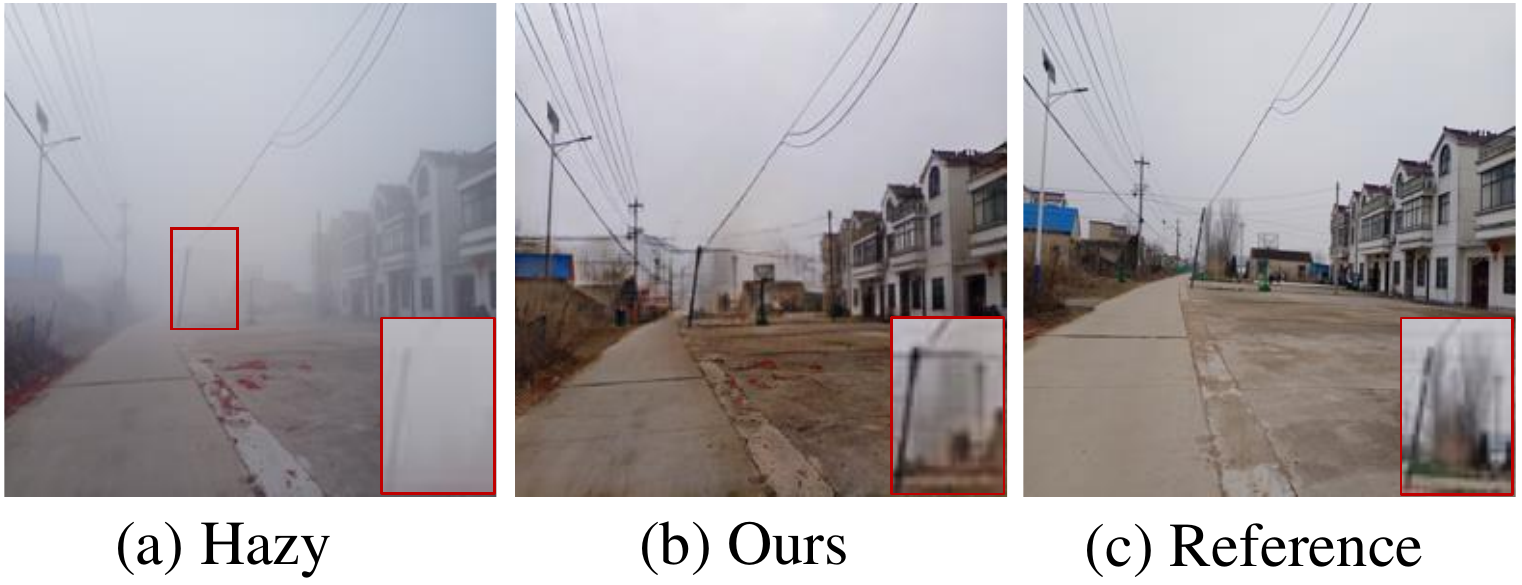}
	\caption{Failures of restoring the objects, which are occluded by dense hazy in the far distance.}
	\label{fig:12}
\end{figure}

\textbf{Limitation.} Here, we discuss the limitation of our non-aligned dehazing framework. Handling dense haze poses a significant challenge as it renders the extraction of meaningful features by the dehazing CNN network difficult, primarily because the input to the network lacks useful information aside from the presence of thick haze. Consequently, our model may occasionally introduce artifacts in the dehazing results. As illustrated in Fig.~\ref{fig:12}, the dehazing outcomes indicated by the red boxes do not meet our desired quality standards and exhibit artifacts. 

\section{Conclusion}
We have presented a novel and effective dehazing framework for real-world images that uses non-aligned supervision. This framework leverages a multi-scale reference loss to compare the prediction of the dehazing network with a clear and non-aligned reference image. It enables the collection of hazy/clear image pairs from real-world environments, even when they are not perfectly aligned. Additionally, our framework includes a mvSA network that uses dark channel prior as location guidance to improve the estimation of the mean and variation of the infinite airlight. The effectiveness of our framework was demonstrated through extensive experiments, which showed that it outperforms state-of-the-art methods in dehazing real-world images.
%In this work, we developed an effectively dehazing framework with non-aligned supervision in real scenes. Specially, we proposed a multi-scale reference loss between the network prediction and the clear and non-aligned image to supervise the dehazing network. This superiority brings the advantage that it is easy to collect the hazy/clear image pair from the real world as the image pair is not strictly aligned. Furthermore, we designed a mvSA network to better predict the mean and variation of the infinite airlight using the dark channel prior. Additionally, we analyzed the effectiveness of the proposed non-aligned supervision dehazing framework. Finally, extensive experiments demonstrated that our framework can effectively dehazing with a favorable performance against state-of-the-art methods.

\backmatter

\bmhead{Supplementary information}

More visual comparison, video demo (only test) and loss curve of train are provided in the \emph{supplementary materials}.

\bmhead{Data availability statement}

Our Phone-Hazy dataset will be made publicly available at \href{https://fanjunkai1.github.io/projectpage/NSDNet/index.html}{project homepage} after the paper is accepted.

\bmhead{Acknowledgments} This work was partially supported by the National Science Fund of China (Grant Nos. 62072242, 62206134).

\begin{appendices}

\section{Phone-Hazy dataset}
\label{Appendix_A:phone-hazydataset}

In pursuit of our goal of gathering hazy/clear image pairs through non-aligned supervision, we have established the Phone-Hazy dataset including 415 pairs. In this subsection, we present its collection procedure and statistical analysis.

\begin{figure}[t]
	\centering
	\includegraphics[width=0.96\linewidth]{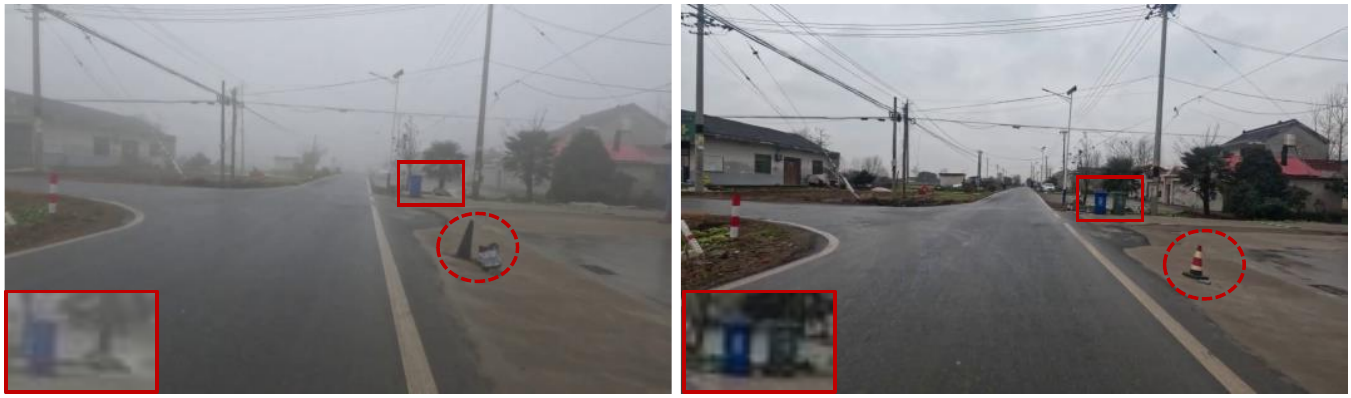}
	\caption{An example of hazy/clear image pairs from the Phone-Hazy dataset, showcasing the contrast between hazy and clear images, revealing noticeable pixel misalignment and semantic inconsistencies.}
	\label{fig:Appendix_A_fig2}
\end{figure}

\textbf{Collection procedure.} The collection scheme is detailed below:
\begin{itemize}
\item 1) Acquire hazy images using an iPhone XR on days with hazy weather conditions, with a preference for focusing on buildings to minimize interference from vehicles and pedestrians.
\item 2) Capture a clean reference image of the same scene is captured under cloudy conditions, using the foggy image as a reference.
\item 3) Perform preprocessing on the hazy/clear image pairs (e.g., cropping) Preprocess the hazy/clear image pairs (\eg, cropping) to maintain their semantic consistency as much as possible, acknowledging that minor inconsistencies are unlikely to have a significant impact.
\end{itemize}
This approach streamlines data collection for real-world scenes. An illustrative example of such an image pair is depicted in Fig.~\ref{fig:Appendix_A_fig2}.

\begin{figure}[t]
	\centering
	\includegraphics[width=1.0\linewidth]{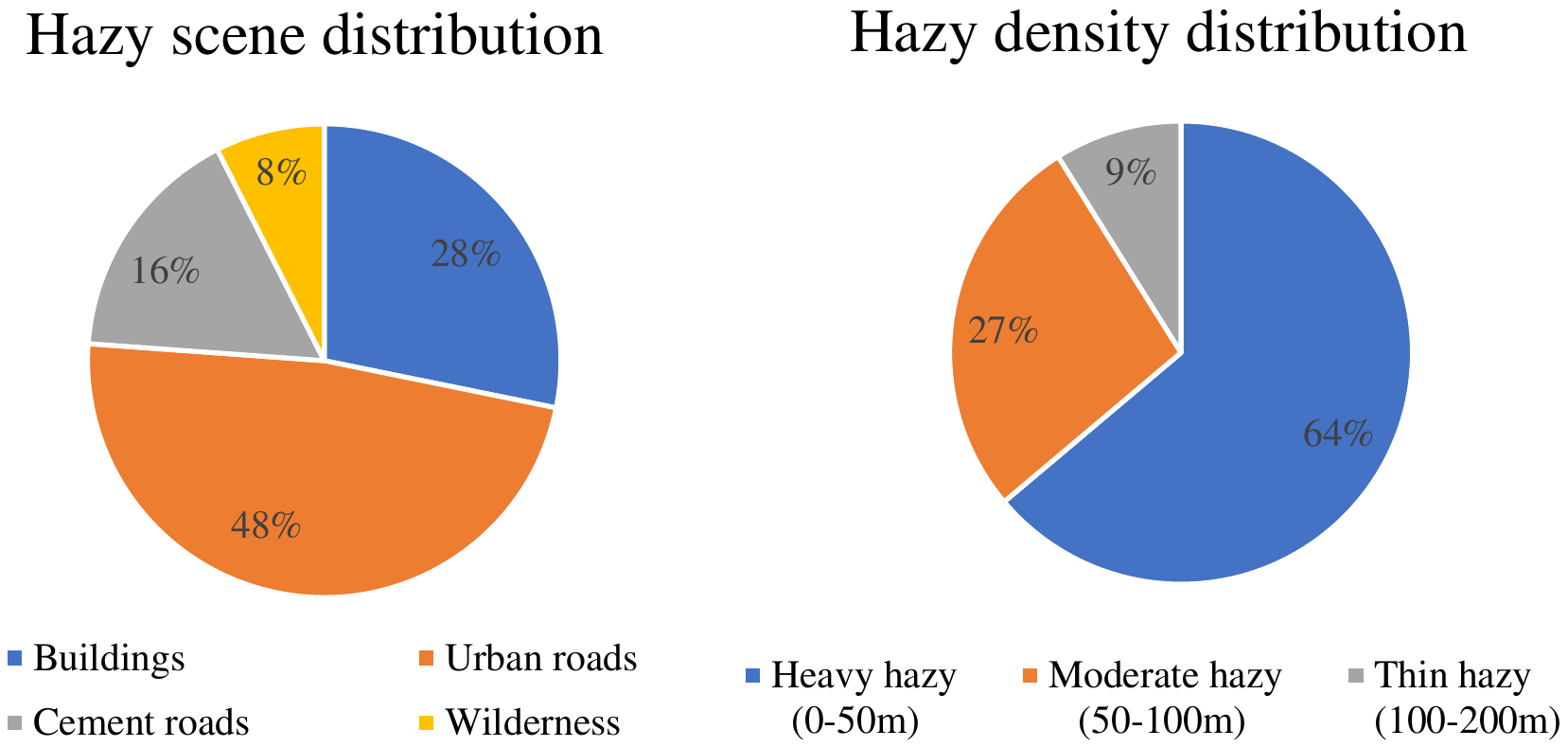}
	\caption{Statistical analysis of hazy scene and density distributions in the Phone-Hazy dataset.}
	\label{fig:Appendix_A_fig1}
\end{figure}

\textbf{Statistical Analysis.} Our primary application context in this work is autonomous driving, with the aim of enhancing the visual perception of autonomous driving systems in hazy weather conditions. As showcased in Fig.~\ref{fig:Appendix_A_fig1}, we have amassed a substantial number of urban road scenes, with the majority of hazy scenes exhibiting a visibility range of 0-50 meters. These scenes are of paramount importance for conducting research aimed at improving visibility in genuine hazy conditions.

In comparison to the 208 image pairs found in the BeDDE dataset~\cite{zhao2019evaluation}, our Phone-Hazy dataset stands out for its larger volume of data and greater scene diversity. Furthermore, the feature similarity between our hazy and reference images is notably higher, as demonstrated in Fig.~\ref{fig:Appendix_A_fig1} and Fig.~\ref{fig:Appendix_A_fig2}. Additionally, while the MRFID dataset~\cite{liu2020image} boasts a substantial volume of data, its scenes predominantly consist of mountains and thus lack the diversity present in Phone-Hazy.

%\noindent {\bf Necessity and advantages of Phone-Hazy.}  \textit{Necessity:} Collecting aligned data for this purpose is a formidable challenge, owing to the substantial variations observed across different scenes, which complicates model training. In light of this, our dataset offers a pragmatic compromise by providing non-aligned data within the same scene. \textit{Advantages:} 1) Phone-Hazy, despite lacking Ground Truth (GT), presents a superior dataset for training models with real-world data. 2) The ease of collecting Phone-Hazy data stands as a notable advantage, with plans to add over 2000 image pairs to the dataset in the near future.

%\noindent {\bf RESIDE dataset\footnote{\url{https://sites.google.com/view/reside-dehaze-datasets/}}.} Following previous works\cite{wu2021contrastive,guo2022image,zheng2023curricular}, we utilize the ITS and OTS subsets of the RESIDE dataset \cite{li2018benchmarking} for training and evaluate on the SOTS subset, consisting of 500 indoor and 500 outdoor hazy images.
%For SOTS dataset, we choose 6000 synthetic hazy images from training, 3000 from ITS and 3000 from OTS. SOTS is used for test, which contains 500 indoor hazy images and 500 outdoor hazy ones. Here, ITS, OTS and SOTS are three subsets of the RESIDE \cite{li2018benchmarking} dataset, respectively.

\section{Results on Synthetic Datasets}\label{Appendix_B}
To further assess the effectiveness of our proposed approach, we conducted an evaluation using the synthetic Foggy Cityscapes dataset~\cite{cordts2016cityscapes}, which can be found at the following link\footnote{\url{https://www.cityscapes-dataset.com/}}. This dataset consists of binocular foggy image pairs. For our research, we focused on the monocular aspect of the dataset, specifically the left image. The training set comprises 8,925 pairs of monocular hazy and clear images, while the validation set includes 1,500 pairs of monocular hazy and clear images.

\begin{figure*}[t]
	\centering
	\includegraphics[width=0.95\linewidth]{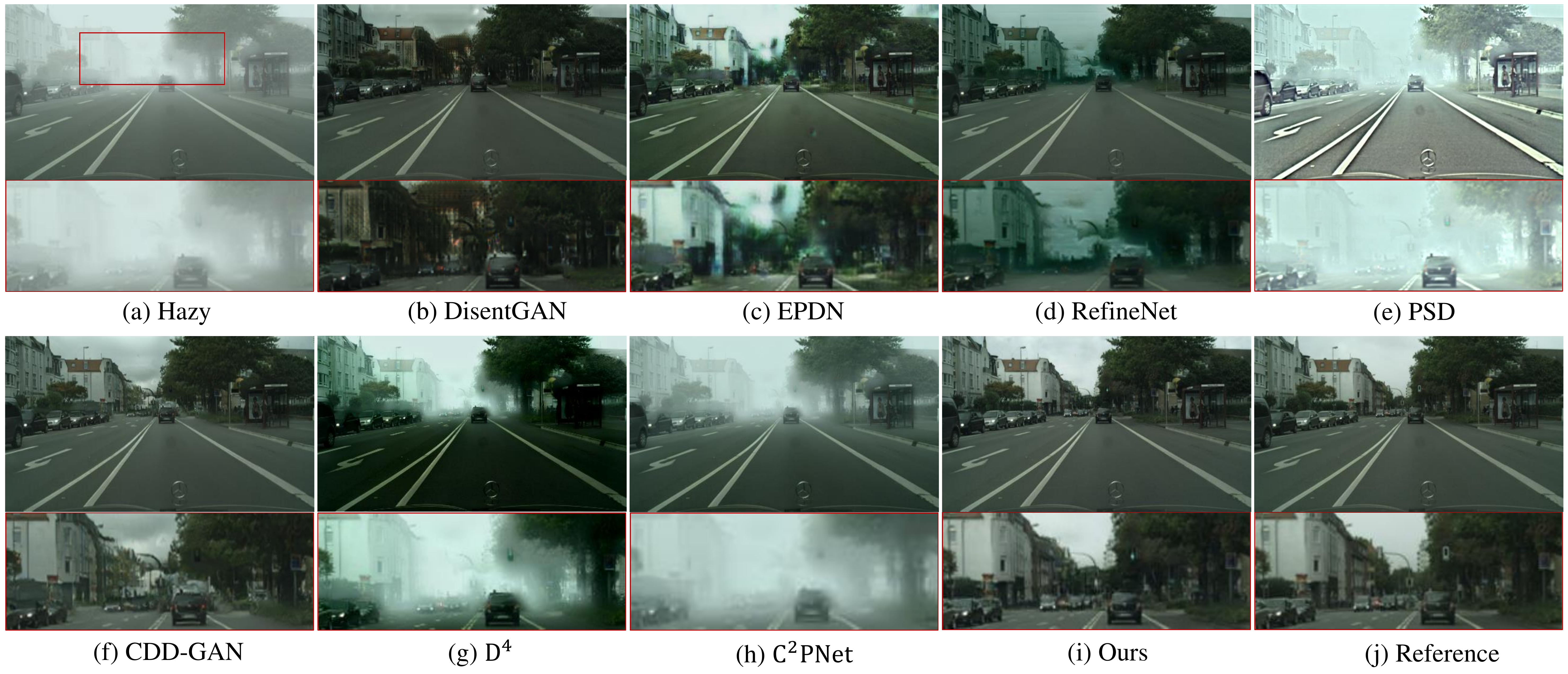}
	\caption{Comparison of dehazing results on synthetic hazy dataset.}
	\label{fig:Appendix_B1}
	\vskip -0.1in
\end{figure*}

Table~\ref{tab:Appendix_B1} presents the evaluation results, including PSNR and SSIM scores. Notably, our proposed method outperforms state-of-the-art techniques when applied to the Foggy Cityscapes dataset, as evidenced by the higher PSNR and SSIM values. To provide a more visual representation of our findings, we offer a set of comparative observations in Fig.~\ref{fig:Appendix_B1}. It is evident that competitive methods exhibit various artifacts and suboptimal detail restoration in their dehazed results. In stark contrast, our proposed method consistently generates significantly clearer results that closely resemble the ground truth images.

%But we did not achieve SOTA results on the SOTS dataset. This is mainly because, in this work, our primary focus was on image dehazing in real-world scenes, and Foggy Cityscapes are closer to real-world haze compared to SOTS. Due to the synthesis of Foggy Cityscapes dataset utilizes atmospheric scattering model and depth map. However, these models (\eg, AECR-Net, DeHamer, C$^{2}$PNet) that ignore atmospheric scattering perform better on the simpler SOTS dataset.

% \emph{More visual results are included in supplementary material}.

\begin{table}[t]
	\linespread{1.0}
	\centering
	\setlength\tabcolsep{1.6pt}%???
	\renewcommand\arraystretch{1.4}
	\scalebox{0.84}{
		\begin{tabular}{c|c|cc|c|c}
			\Xhline{1.3pt}
			\multicolumn{1}{c|}{\multirow{2}[1]{*}{\makecell{Data \\Settings}}}& \multicolumn{1}{c|}{\multirow{2}[1]{*}{Methods}} & \multicolumn{2}{c|}{Foggy Cityscapes} &\multirow{2}[1]{*}{\makecell{Params\\(M)}} &\multirow{2}[1]{*}{Reference}\\
			& & PSNR $\uparrow$  & SSIM $\uparrow$ &  & \\
			\Xhline{0.7pt}
			\multirow{4}[6]{*}{Unpaired} & DCP \cite{he2010single}  & 16.51* & 0.73* & - & CVPR'09 \\
			%Cycle-Dehaze (CVPRW'18) \cite{engin2018cycle} & 21.96* & 0.86*  & 23.37* & 0.92* \\
			& DisentGAN \cite{yang2018towards}   & 25.49* & 0.92* & 11.48 & AAAI'18\\
			& RefineNet \cite{zhao2021refinednet}  & 23.54* & 0.93* & 11.38 & TIP'21 \\
			& CDD-GAN \cite{chen2022unpaired}    & 24.72* & 0.90*  & 29.27 & ECCV'22\\
			& D$^{4}$ \cite{yang2022self}   & 15.16* & 0.62*  & 10.70 & CVPR'22\\
			\Xhline{0.7pt}
			\multirow{7}[6]{*}{Paired} & MSCNN \cite{ren2016single}& 18.99 & 0.86 & 0.008 & ECCV'16 \\
			& DehazeNet \cite{cai2016dehazenet}    & 14.97 & 0.49 & 0.009 & TIP'16 \\
			& AOD-Net \cite{li2017aod}   & 15.45 & 0.63 & 0.002 & ICCV'17 \\
			& FFANet \cite{qin2020ffa}   & 24.54 & 0.94 & 4.456 & AAAI'20 \\
			& EPDN \cite{qu2019enhanced}     & 25.23* & 0.94* & 17.38 & CVPR'19 \\
			& AECR-Net \cite{wu2021contrastive}   & -  &-  & 2.611 & CVPR'21 \\
			& PSD \cite{chen2021psd}   & 11.16* & 0.55* & 33.11 & CVPR'21 \\
			& PMNet \cite{ye2021perceiving}   & -  & -  & -   & ECCV'22 \\
			& Dehamer \cite{guo2022image}  & 26.18*   & 0.93*  & 132.45 & CVPR'22 \\
			& C$^{2}$PNet \cite{zheng2023curricular}& 17.79 & 0.85 &7.71 & CVPR'23 \\
			\Xhline{0.7pt}
			\rowcolor{gray!10}Non-aligned &{\bf NSDNet} (Ours)  & {\bf 29.35} & {\bf 0.96} & 11.38  & - \\
			\Xhline{1.3pt}
	\end{tabular}}%
	\caption{Quantitative comparison of the dehazing results on Foggy Cityscapes dataset. $\uparrow$ represents the higher the better. The symbol "*" means that we retrain the model on the Foggy Cityscapes train set. "-" indicates no training code provided.}
	\label{tab:Appendix_B1}%
\end{table}%

\section{More Methods on RTTS Dataset}\label{Appendix_C}

\begin{table}[t]
	\centering
	%\vskip -0.1in
	\setlength\tabcolsep{1.6pt}%调列距
	\renewcommand\arraystretch{1.4}
	\scalebox{0.86}{
		\begin{tabular}{l|c|c|c|c}
			\Xhline{1.3pt}
			\multicolumn{1}{c|}{\multirow{2}[2]{*}{Data Setting}} & \multirow{2}[2]{*}{Methods} & \multicolumn{2}{c|}{RTTS (Only testing)} & \multirow{2}[1]{*}{Reference} \\
			&   & \multicolumn{1}{c}{FADE $\downarrow$} & \multicolumn{1}{c|}{NIQE $\downarrow$} & \\
			\Xhline{0.7pt}
			\multicolumn{1}{c|}{\multirow{6}[2]{*}{\makecell{Aligned \\(\ie, need GT)}}} & MSBDN \cite{dong2020multi} & 1.3952  & 4.5269 & CVPR'20 \\
			& FFANet \cite{qin2020ffa}   & 2.0734  & 4.9281 & AAAI'20 \\
			& UHD \cite{zheng2021ultra}   & 1.2457  & 4.4244 & CVPR'21 \\
			& IPUDN  \cite{kar2020transmission} & 0.9126  & 7.3470  & Arxiv'22 \\
			& PMNet \cite{ye2021perceiving} & 1.4211  & 4.7523 & ECCV'22 \\
			& DeHamer \cite{guo2022image} & 1.3062  & 7.2924 & CVPR'22 \\
			& C$^{2}$PNet \cite{zheng2023curricular} & 2.0616  & 5.0383  & CVPR'23 \\
			\Xhline{0.7pt}
			\rowcolor{gray!10}Non-aligned & \textbf{NSDNet} (Ours) & \textbf{0.7419}  & \textbf{3.6905} & - \\
			\Xhline{1.3pt}
	\end{tabular}}%
	\caption{Comparison of the proposed method and methods with aligned ground truth on RTTS dataset.}
	\label{tab:Appendix_C1}%
\end{table}%

\begin{figure*}[t]
	\centering
	\includegraphics[width=0.95\linewidth]{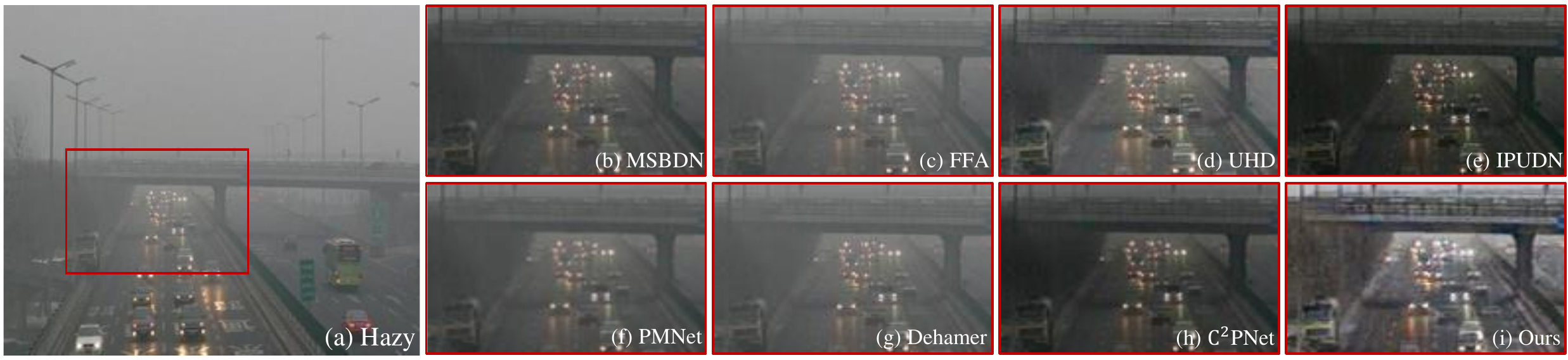}
	\caption{Visual comparison of dehazing results from models heavily dependent on ground truth (GT) on the RTTS dataset.}
	\label{fig:Appendix_C_fig1}
\end{figure*}

All dehazing models aim to address real-world haze removal challenges. For models that heavily rely on ground truth (GT) training data, we conduct a fair evaluation of their effectiveness using exclusively real-world hazy datasets. Quantitative and visual comparisons are presented separately in Table \ref{tab:Appendix_C1} and Fig.~\ref{fig:Appendix_C_fig1}.  However, it is evident from the visual results that these models, trained initially on synthetic data and then adapted to real-world scenes, do not yield satisfactory dehazing outcomes.

\section{More Visual Results Using our NSDNet}\label{Appendix_D}

\begin{figure*}[t]
	\centering
	\includegraphics[width=0.96\linewidth]{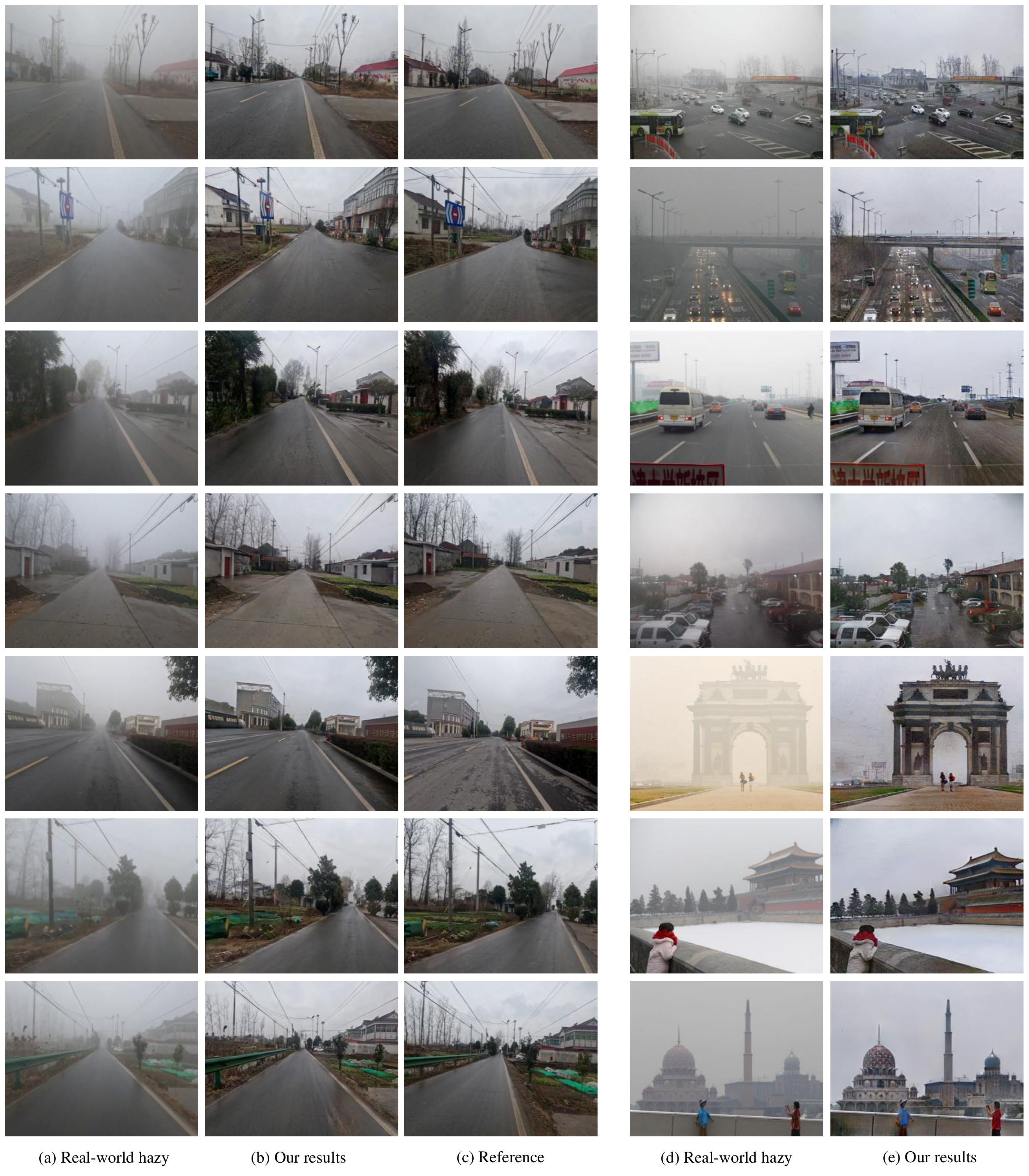}
	\caption{More visual results. (a)-(c) showcase visual outcomes produced by our NSDNet using the Phone-Hazy dataset, whereas (d) - (e) present visual representations of our model's performance on the RTTS dataset. These two real-world datasets collectively highlight the efficacy of our model.}
	\label{fig:Appendix_D_fig1}
\end{figure*}

\textbf{More Visual Results on Phone-Hazy.} In order to demonstrate the effectiveness of our real-world dehazing model, NSDNet, in the context of autonomous driving scenarios, we have curated additional urban road scenes from the Phone-Hazy dataset. These scenes serve as validation for our approach, and the corresponding results are presented visually in Fig.~\ref{fig:Appendix_D_fig1}(a)-(c).

\textbf{More Visual Results on RTTS.} To underscore the robustness of our model, we have also handpicked challenging hazy images from the RTTS dataset. These images depict complex scenarios featuring people and vehicles in heavy haze, as well as long-range hazy conditions. We employ these images to assess the performance of our real-world dehazing model and provide a comparative visualization in Fig.~\ref{fig:Appendix_D_fig1}(d)-(e).

\end{appendices}

\clearpage

%%===========================================================================================%%
%% If you are submitting to one of the Nature Portfolio journals, using the eJP submission   %%
%% system, please include the references within the manuscript file itself. You may do this  %%
%% by copying the reference list from your .bbl file, paste it into the main manuscript .tex %%
%% file, and delete the associated \verb+\bibliography+ commands.                            %%
%%===========================================================================================%%

\bibliographystyle{sn-mathphys} 
\bibliography{sn-bibliography}% common bib file
%% if required, the content of .bbl file can be included here once bbl is generated
%%\input sn-article.bbl

\end{document}